\documentclass[acmtog]{acmart}
\acmSubmissionID{1387}

\usepackage{placeins}
\usepackage{listings}
\usepackage{soul}
\usepackage{multirow}
\hypersetup{
    colorlinks=true,
    linkcolor=blue,
    filecolor=magenta,      
    urlcolor=cyan,
}

\usepackage{colortbl}
\usepackage{hhline}

\definecolor{lightgrey}{RGB}{246,246,246}
\definecolor{lightblue}{RGB}{240,245,255}
\definecolor{darkblue}{RGB}{40,40,85}
\definecolor{turquoise}{cmyk}{0.65,0,0.1,0.3}
\definecolor{blue}{rgb}{0.0,0.1,0.9}
\definecolor{purple}{rgb}{0.6,0.2,0.6}
\definecolor{dark_green}{rgb}{0, 0.5, 0}
\definecolor{orange}{rgb}{0.8, 0.6, 0.2}
\definecolor{red}{rgb}{0.8, 0.2, 0.2}
\definecolor{light_grey}{rgb}{0.95, 0.95, 0.95}
\definecolor{white}{rgb}{1.0, 1.0, 1.0}
\definecolor{chtholly_blue}{HTML}{0072C3}

\newif\ifcolorcodedcomments
\colorcodedcommentsfalse

\ifcolorcodedcomments
\newcommand{\TODO}[1]{{\color{red}[TODO: #1]}}

\newcommand{\TTComments}[1]{{\color{turquoise}[TT: #1]}}

\newcommand{\MRComments}[1]{{\color{violet}[MR: #1]}}
\newcommand{\YTComments}[1]{{\color{brown}[YT: #1]}}
\else
\newcommand{\TODO}[1]{}

\newcommand{\TTComments}[1]{}

\newcommand{\MRComments}[1]{}
\newcommand{\YTComments}[1]{}
\fi

\begin{document}

\setcopyright{cc}
\setcctype{by}
\acmJournal{TOG}
\acmYear{2026}
\acmVolume{45}
\acmNumber{6}
\acmArticle{181}
\acmMonth{12}
\acmDOI{10.1145/3842521}

\title{
    \raisebox{-0.25em}{\includegraphics[height=1.1em]{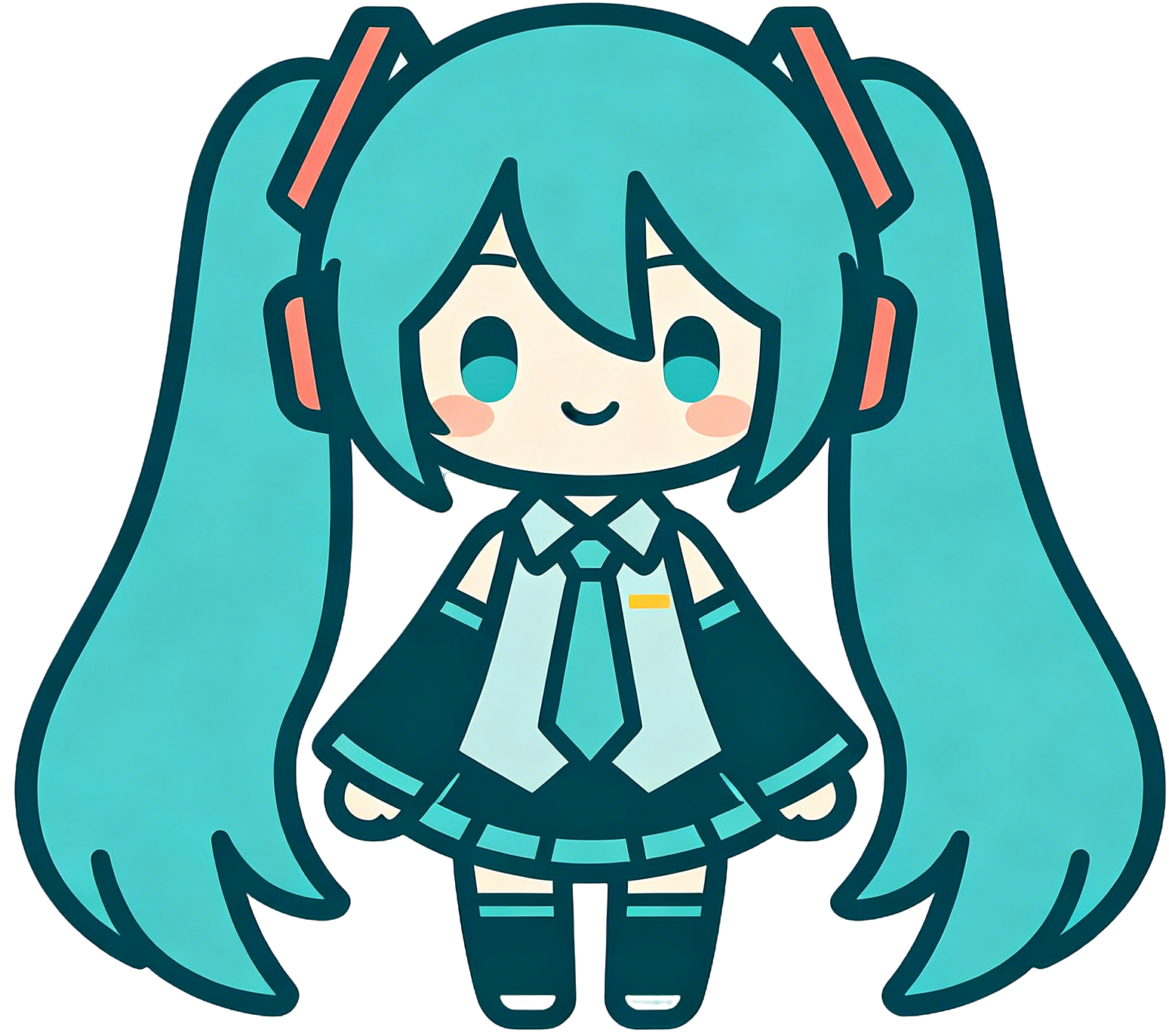}}
    \enskip
    \textbf{MIKU}:
    \textbf{M}aterial
    \textbf{I}nvariance and
    \textbf{K}nowledge
    \textbf{U}nification for PBR Generation
}

\author{Yajie Bao}
\authornote{This work was conducted during an internship at DreamTech.}
\orcid{0009-0007-6497-7557}
\affiliation{%
  \institution{Nanjing University, China and DreamTech}
  \country{China}
}
\email{652025710001@smail.nju.edu.cn}

\author{Jiaye Fu}
\orcid{0009-0008-3722-811X}
\affiliation{%
  \institution{Peking University, China and DreamTech}
  \country{China}
}
\email{jyfu@stu.pku.edu.cn}

\author{Youtian Lin}
\orcid{0000-0002-9179-9406}
\affiliation{%
  \institution{Nanjing University}
  \country{China}
}
\email{linyoutian.loyot@gmail.com}

\author{Shuang Wu}
\orcid{0009-0002-2939-4448}
\affiliation{%
  \institution{Nanjing University, China and DreamTech}
  \country{China}
}
\email{wushuang@smail.nju.edu.cn}

\author{Jiachen Qian}
\orcid{0009-0009-0538-5433}
\affiliation{%
  \institution{The University of Hong Kong, Hong Kong and DreamTech}
  \country{China}
}
\email{qianjiachen@dreamtech.ai}

\author{Feihu Zhang}
\orcid{0000-0003-0712-5722}
\affiliation{%
  \institution{DreamTech}
  \country{China}
}
\email{zhangfeihu@dreamtech.ai}

\author{Xun Cao}
\orcid{0000-0003-3094-4371}
\affiliation{%
  \institution{Nanjing University}
  \country{China}
}
\email{caoxun@nju.edu.cn}

\author{Yao Yao}
\authornote{Corresponding author.}
\orcid{0000-0001-9866-4291}
\affiliation{%
  \institution{Nanjing University}
  \country{China}
}
\email{yaoyao@nju.edu.cn}



\definecolor{revisionblue}{RGB}{0,80,200}

\newif\ifshowrevision
\showrevisionfalse

\newcommand{\rev}[1]{%
  \ifshowrevision
    \textcolor{revisionblue}{#1}%
  \else
    #1%
  \fi
}

\newenvironment{revision}
  {\ifshowrevision\color{revisionblue}\fi}
  {}

\begin{abstract}
In physically based rendering (PBR) content creation, high-quality material generation remains a fundamental challenge, primarily hindered by the absence of standardized representations across existing datasets and synthesis pipelines. The heterogeneity in material definitions, rendering conventions, and parameterization schemes introduces significant inconsistencies, ultimately leading to unstable and physically implausible PBR synthesis results. To systematically address these limitations, 
\begin{revision}we propose MIKU, a unified framework that establishes standardized PBR asset representation and enables physically consistent material generation across diverse 3D content domains.\end{revision}
Our core contribution is a comprehensive unified PBR material processing pipeline that standardizes the entire workflow from data preprocessing to final texture synthesis. This pipeline enforces consistent representation across geometry and texture domains throughout material generation. Critically, it addresses the long-standing challenge of inconsistent material parameterizations by introducing a normalization scheme that harmonizes diverse input sources into a common representation space, enabling robust cross-dataset training and inference.Building upon this principled formulation, we train a \rev{novel physically aware diffusion model} specifically designed for PBR texture generation. Our model operates within a two-stage multi-view synthesis framework.The first stage focuses on geometry-aware albedo generation, where multi-view albedo maps are synthesized under explicit geometric guidance to ensure cross-view appearance consistency. The second stage performs metallic--roughness (MR) generation using our trained diffusion model, which predicts physically plausible roughness and metallic maps conditioned on the generated albedo and the unified representation. The model is trained end-to-end on standardized data, learning to generate materials that \rev{respect category-level material priors and semantic part boundaries}.By explicitly decoupling representation standardization from appearance synthesis, MIKU substantially improves both the stability and physical consistency of generated materials. Extensive experiments demonstrate that our approach achieves state-of-the-art performance in both texture quality and semantic consistency across different object parts, significantly advancing automated material generation for 3D content creation workflows.
\end{abstract}


\begin{teaserfigure}
  \includegraphics[width=\textwidth]{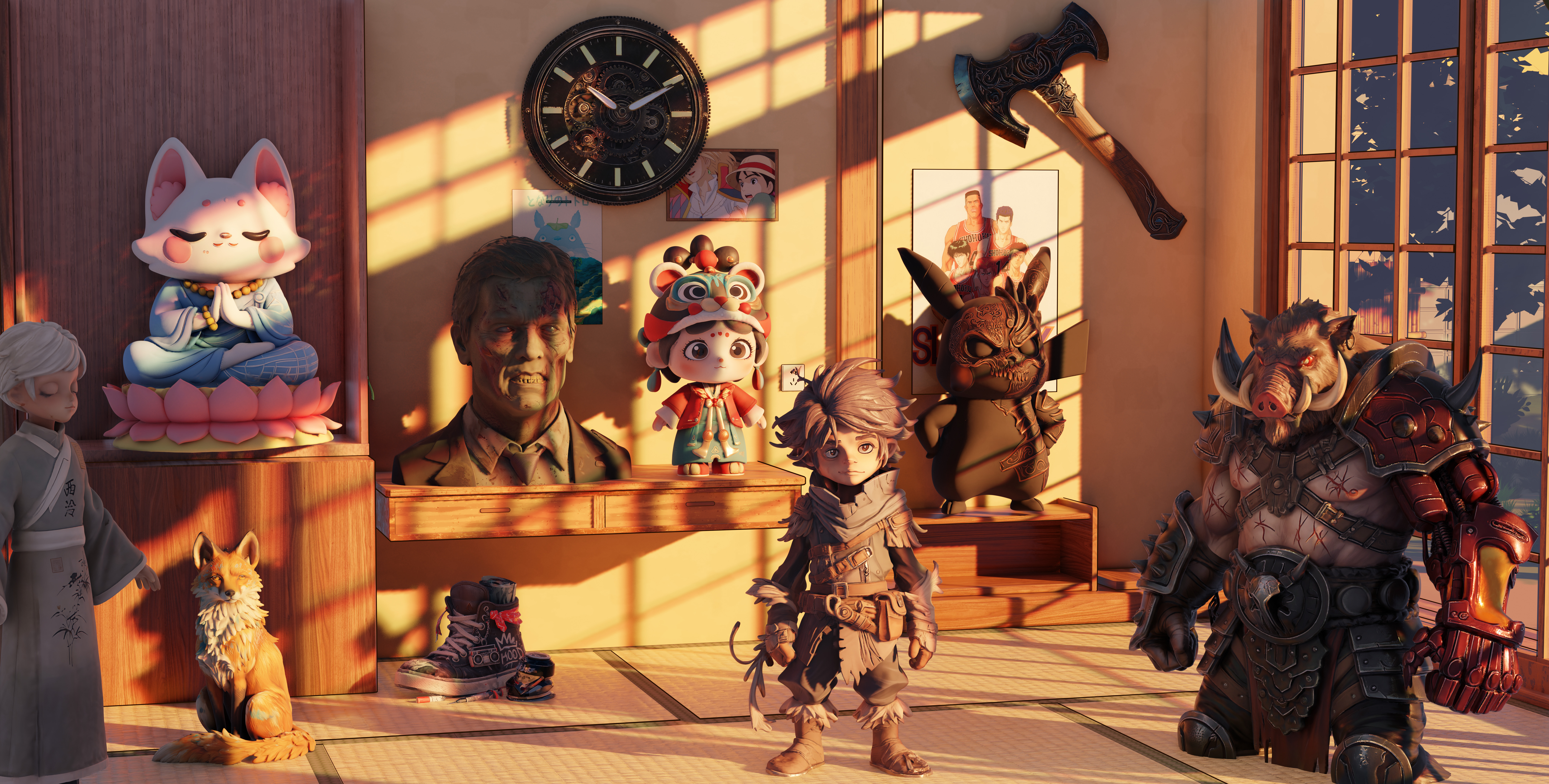}
  \caption{The entire scene is assembled from AIGC-generated meshes, where both the surface textures and PBR material maps are consistently synthesized by our method.}
  \Description{A scene assembled from generated 3D meshes, showing consistent textures and physically based material maps under varied illumination.}
  \label{fig:teaser}
\end{teaserfigure}

\maketitle

\begin{figure*}[t]
    \centering
    \includegraphics[width=\textwidth]{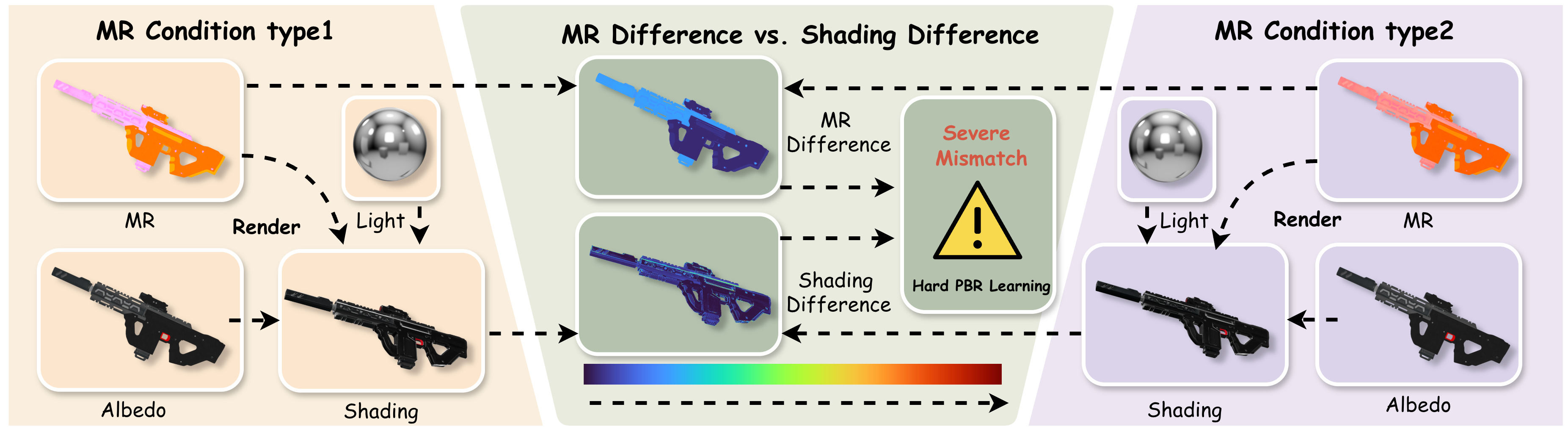}
    \caption{Illustration of the discrepancy attenuation in image-supervised PBR learning. Even with fixed geometry, albedo, camera view, and illumination, substantially different MR conditions may result in only minor shading differences after rendering. }
    \label{fig:pbr_ambiguity}
\end{figure*}
\section{Introduction}
The generation of high-fidelity textures for 3D models is a fundamental component of realistic digital asset creation and has attracted sustained research interest in the graphics and vision communities \cite{poole2022dreamfusion, shi2023mvdream, liu2023syncdreamer, jin2024mv, liu2024text, tang2024mvdiffusion++, he2025materialmvp}. Among texture generation methods, physically based rendering (PBR) has become a central paradigm for producing realistic materials by modeling surface reflectance with properties such as albedo, roughness, and metallic, enabling consistent rendering under varying illumination. This has motivated increasing efforts in PBR-aware texture generation.

A line of work adopts single-view image-conditioned formulations, where generation is guided by a single reference view to transfer appearance cues such as structure and material properties to the output. Methods such as IntrinsicAnything~\cite{chen2024intrinsicanything}, IntrinsiX~\cite{kocsis2025intrinsix}, DualMat~\cite{huang2025dualmatpbrmaterialestimation}, and RGB$\leftrightarrow$X~\cite{Zeng_2024} leverage diffusion priors, cross-intrinsic attention, and bidirectional or dual-path diffusion frameworks to enhance intrinsic decomposition and improve cross-channel consistency. However, these approaches are fundamentally constrained by single-view ambiguity, which often leads to incomplete geometric reasoning and limited cross-view consistency.

Another line of work adopts multi-view image-conditioned formulations, where generation is guided by a set of calibrated or uncalibrated observations to better constrain 3D appearance reasoning. Methods such as SuperMat~\cite{hong2025supermat}, MVPainter~\cite{shao2025mvpainter}, IDArb~\cite{li2024idarb}, MaterialMVP~\cite{he2025materialmvp}, MuMA~\cite{zhu2025muma3dpbrtexturing}, PacTure~\cite{fei2025pacture}, PBR3DGen~\cite{wei2026pbr3dgen}, MCMat~\cite{zhu2024mcmat}, and LumiTex~\cite{bao2025lumitex} build upon diffusion-based architectures with cross-view and cross-domain attention, geometry or reference-aware conditioning, and multi-view feature aggregation to jointly synthesize physically based material properties with improved spatial coherence and appearance consistency across views. Despite these advantages, we observe that the training process of models for PBR attribute prediction is often unstable, particularly under varying data distributions and view configurations.  In addition, the learned representations often exhibit limited consistency in PBR material estimation, where visually similar or semantically identical objects may be assigned different predictions of physically based properties across instances or views. Furthermore, their generalization capability is largely restricted to specific training datasets and predefined data distributions, resulting in degraded robustness when encountering unseen categories, diverse material patterns, or in-the-wild 3D assets.

This raises a central question: \textbf{Does the observed inconsistency primarily stem from limitations of the generative model, or from the inherent ambiguity in PBR material definitions?} Addressing this question is non-trivial, as learning PBR material attributes from image observations is intrinsically ill-posed. A fundamental challenge in learning PBR material attributes is the ambiguity of artist-defined reflectance annotations. For the same object, different artists may assign substantially different metallic--roughness (MR) configurations, all of which can be physically plausible yet correspond to distinct material interpretations and visually reasonable renderings. Moreover, even highly similar rendered appearances may originate from markedly different MR definitions, indicating that small image-space variations do not necessarily imply small discrepancies in the underlying material-parameter space. As shown in Fig.~\ref{fig:pbr_ambiguity}, even when geometry, albedo, camera view, and illumination are fixed, large discrepancies in MR space can be compressed into only subtle differences in the rendered shading. This weakens image-space supervision and makes material recovery an under-constrained inverse rendering problem, where multiple PBR configurations can explain similar appearances. Training on such ambiguous annotations encourages networks to regress toward averaged or over-smoothed material estimates, leading to reduced material contrast, loss of high-frequency reflectance details, and inconsistent predictions across views or instances.

Motivated by the above observations, we introduce a unified data construction and learning framework for physically consistent PBR material generation. Specifically, for each artist-authored asset, we establish a compact PBR material taxonomy and render the asset from six calibrated viewpoints. We filter low-quality samples using a joint complexity metric over albedo and material (MR) variance. The remaining assets are processed by a vision-language model for material recognition, followed by SAM3~\cite{carion2025sam}-based segmentation guided by predicted material prompts. We further apply affine projection with residual injection and manual refinement to preserve fine-grained details and ensure cross-view consistency. On the resulting dataset, we train a two-stage packed PBR generator under a multi-resolution and multi-view formulation, enabling physically consistent material synthesis. The main contributions of this work are summarized as follows.:
\begin{itemize}
    \item We propose a unified data processing pipeline for constructing high-quality PBR datasets, which enforces consistent material definitions, controlled illumination, and semantic alignment across assets. This pipeline produces large-scale, standardized training data with improved diversity and reliability.

    \item We introduce an end-to-end multi-view PBR prediction model that operates at $1024 \times 1024$ resolution and achieves inference within one minute. The model jointly predicts geometry-consistent material properties across views, enabling efficient and high-fidelity PBR generation.

    \item We identify and demonstrate that inconsistency in material definitions and illumination introduces significant ambiguity into the supervision signal, which negatively impacts model stability. By enforcing data-level consistency, our approach reduces this ambiguity and leads to more stable and coherent material predictions across spatial regions and viewpoints.
\end{itemize}

\section{Related Work}
Recent 3D generation methods have shifted from geometric modeling alone to jointly synthesizing geometry and physically-based rendering (PBR) materials, requiring disentangled intrinsic channels—albedo, roughness, and metallic—that remain consistent across views; however, prior methods suffer from entanglement with lighting and inconsistent artist-defined annotations, which degrade relighting, destabilize training, and increase variance across object categories, motivating our approach, which addresses both issues.
\paragraph{\textbf{Texture Generation}}
Existing texture generation methods can be broadly grouped into several categories according to their model architectures, input representations, and supervision settings.
The first category builds upon large-scale pretrained image diffusion models~\cite{podell2023sdxl,peebles2023scalable,flux-2-2025,rombach2022high,flux2024,ramesh2021zero}.
An important line of work leverages Score Distillation Sampling (SDS)~\cite{poole2022dreamfusion}, which optimizes 3D representations by distilling gradients from pretrained 2D diffusion models, providing implicit multi-view supervision. Subsequent works improve geometry fidelity, view consistency, and material quality through enhanced regularization, guidance strategies, and hybrid representations~\cite{lin2023magic3dhighresolutiontextto3dcontent,chen2023fantasia3ddisentanglinggeometryappearance,wang2023prolificdreamerhighfidelitydiversetextto3d,tang2023dreamgaussian}.
Building upon these advances, another branch explicitly synthesizes view-consistent multi-view observations by incorporating geometric guidance (e.g., depth, normals, or camera poses) to enforce cross-view alignment~\cite{long2024wonder3d,huang2025mv,liang2025unitex,he2025materialmvp,hong2025supermat,lai2025hunyuan3d,zeng2024paint3d,bao2025lumitex,liu2025calitex,liu2024text,li2025step1x,hunyuan3d2025hunyuan3d}. The second category formulates multi-view generation within a video diffusion framework, where different viewpoints are modeled as temporally correlated frames. Temporal attention and motion-aware priors are leveraged to improve cross-view consistency and reduce inter-view discrepancies~\cite{voleti2024sv3d,yuan2025seqtex,zuo2024videomv,kim2025rapidmv}. The third category explores native 3D generation methods~\cite{zeng2025textrix,lai2025natex,chen2025lafite,yu2024texgen,xiang2025native}, which explicitly model 3D representations and generate texture through rendering-based or 3D-aware generative pipelines, thereby enforcing geometric consistency in a unified spatial domain. Complementary to texture generation, mesh generation methods~\cite{wu2024direct3d,wu2025direct3d,lai2025hunyuan3d,jia2025ultrashape,xiang2025structured} synthesize textured 3D assets by first generating geometry and then producing textures via UV mapping or image-based refinement within an AIGC pipeline.

\paragraph{\textbf{PBR Generation}}
Physically-based rendering (PBR) generation aims to infer intrinsic material properties, such as albedo, metallic, and roughness, that are consistent with underlying geometry and illumination, enabling physically faithful rendering of 3D assets. \begin{revision}
The first category focuses on single-view PBR generation, which estimates material properties from a single image under limited geometric and lighting cues, often suffering from inherent ambiguity due to under-constrained observations~\cite{kocsis2025intrinsix,chen2024intrinsicanything,luo2025matpedia,huang2025dualmatpbrmaterialestimation,Zeng_2024,xue2026videoneumatneuralmaterialextraction,xue2024reflectancefusion,raghavan2025generative,ying2025chord,vecchio2024matfuse}. The second category explores multi-view image-based PBR generation, which leverages multiple observations to learn view-consistent material representations and improves texture-level synthesis through cross-view supervision~\cite{li2024idarb,he2025materialmvp,zhu2025muma3dpbrtexturing,wei2026pbr3dgen,zhu2024mcmat,bao2025lumitex,munkberg2025videomat,hasselgren2026videomatgen}.\end{revision}

\begin{figure}[t] 
  \centering
  \includegraphics[width=\linewidth]{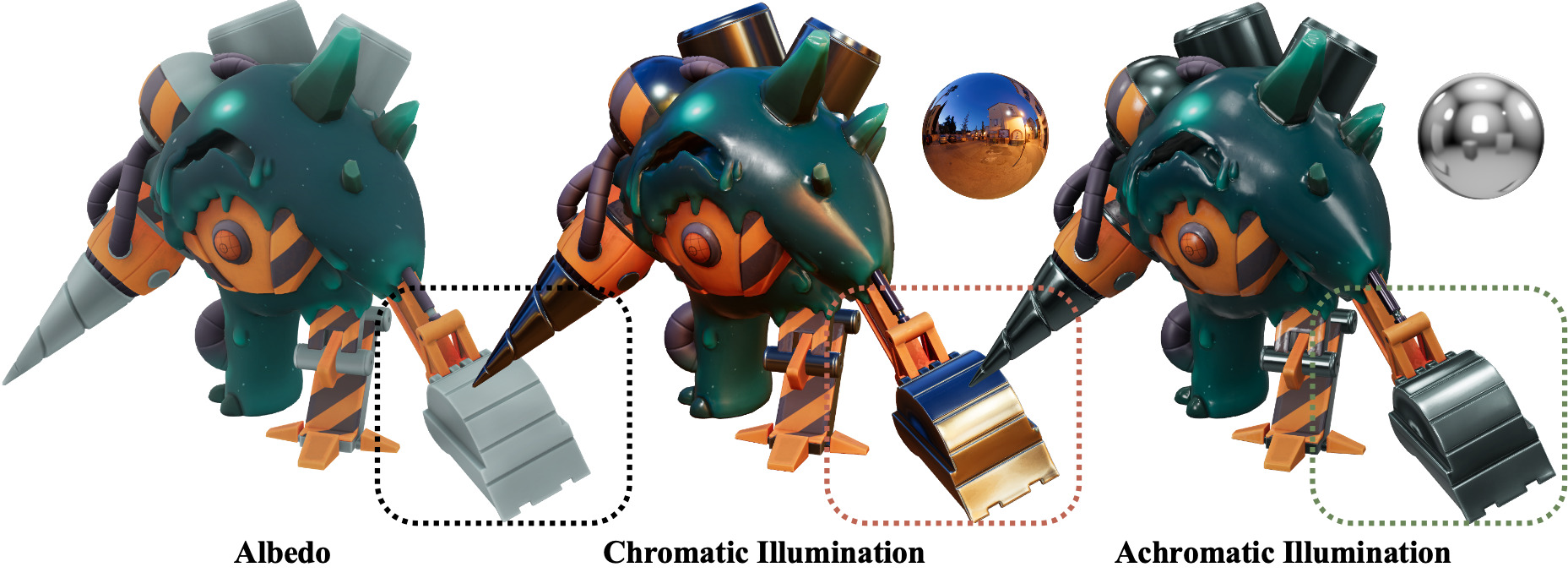} 
  \caption{
Visualization of our color-preserving illumination strategy. Compared with chromatic illumination, which introduces color casts and appearance ambiguity, achromatic illumination preserves intrinsic albedo colors while providing shading cues. 
  }
  \Description{A pipeline illustrating deterministic diffusion-based multi-view PBR material estimation with packed inputs, conditional guidance, and LoRA fine-tuning.}
    \label{fig:render_setting}
\end{figure}

\begin{figure*}[ht]
  \centering
  \includegraphics[width=\textwidth]{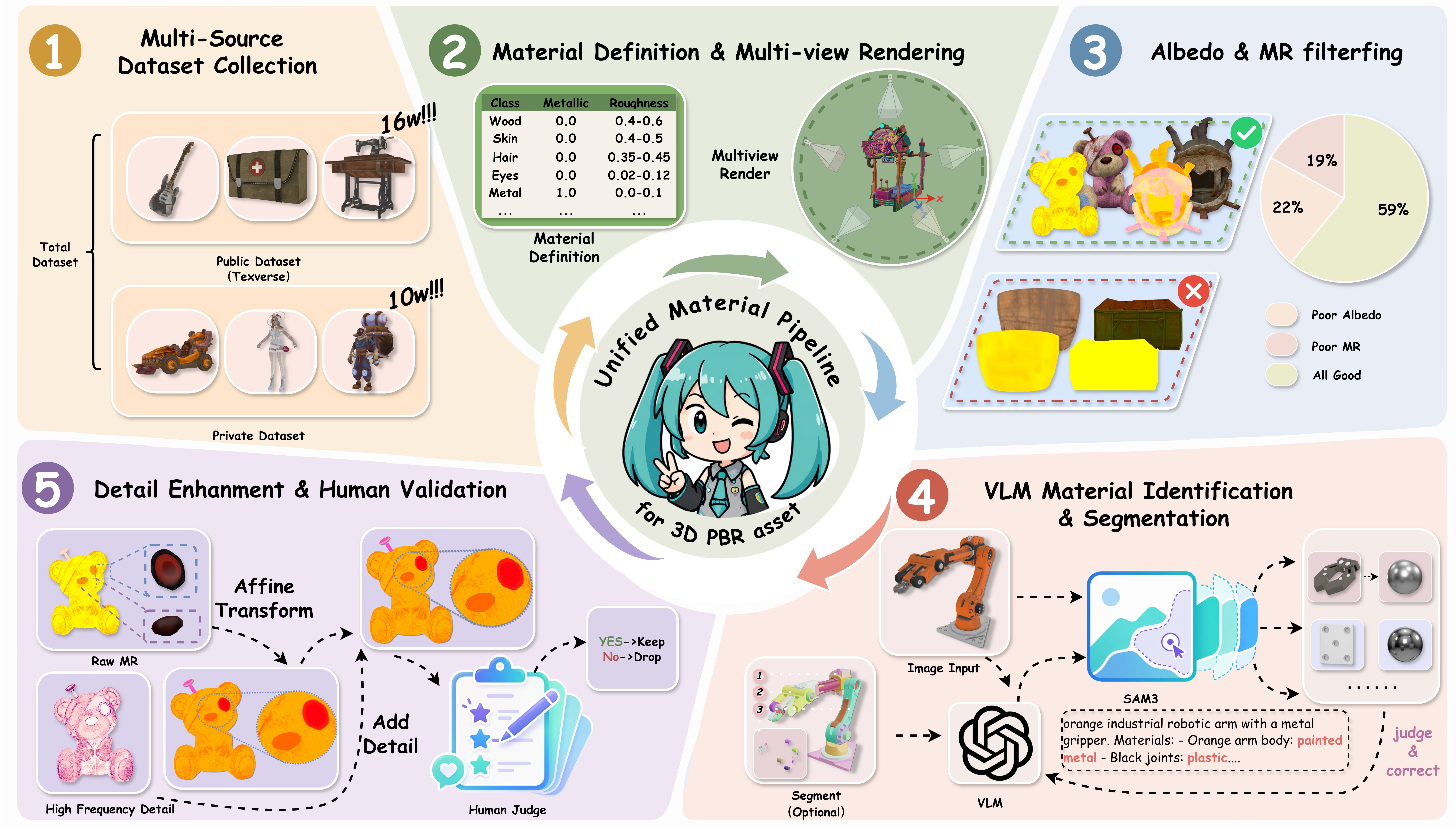}
  \caption{Overview of Our Multi-Stage Pipeline for High-Quality PBR Material Dataset Construction and Refinement. This figure illustrates our end-to-end pipeline for constructing a high-quality PBR material dataset. We first collect multi-source data from the public TexVerse dataset and private asset collections. We then define physically based material properties and render multi-view observations under controlled settings. Next, we apply albedo extraction and PBR-aware filtering to remove low-quality samples. Vision-Language Models (VLMs) with SAM3 are used for material recognition and semantic segmentation to enable fine-grained decomposition. Finally, we enhance texture details via affine transformations and high-frequency boosting, followed by human validation to ensure data quality.}
  \Description{A pipeline illustrating deterministic diffusion-based multi-view PBR material estimation with packed inputs, conditional guidance, and LoRA fine-tuning.}
  \label{fig:data_process}
\end{figure*}

\begin{table}[t]
\caption{\rev {Metallic--roughness (MR) definitions for the 20 predefined semantic material categories and the four optional fine-grained metal subclasses under our unified PBR specification. Each category is associated with a representative MR range that serves as a consistent material prior during data standardization.}}
\label{tab:material_taxonomy}
\centering
\scriptsize
\setlength{\tabcolsep}{3pt}

\ifshowrevision
    \color{revisionblue}
\fi

\begin{tabular}{@{}p{0.52\columnwidth}p{0.18\columnwidth}p{0.20\columnwidth}@{}}
\toprule
\textbf{Material Class} &
\textbf{Metallic} &
\textbf{Roughness} \\
\midrule
Skin                & 0.0 & 0.40--0.50 \\
Eye                 & 0.0 & 0.02--0.12 \\
Hair                & 0.0 & 0.35--0.45 \\
Metal               & 1.0 & 0.00--0.10 \\
Teeth               & 0.0 & 0.28--0.34 \\
Lips                & 0.0 & 0.30--0.36 \\
Plastic             & 0.0 & 0.32--0.38 \\
Rubber              & 0.0 & 0.48--0.61 \\
Glass               & 0.0 & 0.02--0.05 \\
Transparent Liquid  & 0.0 & 0.01--0.04 \\
Ceramic             & 0.0 & 0.15--0.26 \\
Stone               & 0.0 & 0.66--0.73 \\
Wood                & 0.0 & 0.40--0.60 \\
Leather             & 0.0 & 0.49--0.56 \\
Fabric              & 0.0 & 0.77--0.84 \\
Fur                 & 0.0 & 0.73--0.80 \\
Paper               & 0.0 & 0.82--0.89 \\
Bone                & 0.0 & 0.47--0.54 \\
Gemstone            & 0.0 & 0.03--0.07 \\
Gold                & 1.0 & 0.09--0.15 \\
\midrule
Mirror Metal        & 1.0 & 0.01--0.03 \\
Polished Metal      & 1.0 & 0.19--0.25 \\
Semi-Rough Metal    & 1.0 & 0.35--0.42 \\
Matte Metal         & 1.0 & 0.68--0.76 \\
\bottomrule
\end{tabular}

\end{table}\textbf{}

\section{Dataset Curation}

\subsection{Relighted Data for Albedo Learning}Environmental illumination plays a critical role in rendering PBR training data. As illustrated in Fig.~\ref{fig:render_setting}, chromatic illumination can introduce strong color casts and spatially varying shading effects, leading to substantial discrepancies between the observed appearance and the underlying albedo. In particular, under chromatic illumination, the reflected appearance is jointly determined by intrinsic material properties and the spectral composition of the environment, which increases the ambiguity of albedo estimation since multiple albedo--illumination combinations may explain similar observations.

To reduce this entanglement, we primarily adopt achromatic illumination, as shown in the third column of Fig.~\ref{fig:render_setting}, where the lighting is constrained to be color-neutral and mainly contributes highlights and shadows without introducing additional chromatic cues. This setting encourages the model to focus on geometry-dependent shading and material-dependent reflectance variations, enabling more stable recovery of intrinsic albedo from shaded observations. It also reduces the risk that the network incorrectly attributes illumination-induced color shifts to the underlying surface reflectance. Nevertheless, real-world inference scenarios may still involve chromatic light sources, such as neon city environments, sunset scenes, or outdoor scenes with strong environmental color casts. Therefore, rather than completely excluding colored lighting, we use achromatic illumination with a probability of 90\% and chromatic illumination with a probability of 10\% during training. This mixed strategy improves robustness to realistic colored lighting while maintaining a predominantly color-preserving supervision signal for PBR material learning.

\subsection{Semantic-Aligned PBR Curation}
As illustrated in Fig.~\ref{fig:data_process}, our dataset construction begins by consolidating assets from two complementary sources: the publicly available TexVerse dataset~\cite{zhang2025texverse}, which contains approximately 160K assets, and a curated in-house collection of 100K high-quality proprietary assets. Before data cleaning, filtering, and material refinement, this initial asset pool comprises approximately 260K PBR assets in total, providing a large-scale starting corpus for subsequent dataset construction.

\textbf{Material Taxonomy and Multi-view Rendering.}

\begin{revision}
For each artist-authored asset, we convert its material annotations into a unified PBR attribute space rather than relying on dataset-specific material definitions. We first organize the materials into 20 predefined semantic categories and map each category to a canonical metallic--roughness (MR) representation, as summarized in Table~\ref{tab:material_taxonomy}. Rather than manually assigning arbitrary parameters or enforcing a single fixed value for each category, we estimate the empirical MR distribution of each material class from the collected artist-authored assets. For each category, we identify the dominant high-density region and adopt the corresponding MR interval as its representative reference range. For example, dielectric materials such as wood are assigned a metallic value of 0 and exhibit a dominant roughness range of approximately 0.4--0.6, whereas highly reflective metallic materials typically have metallic values close to 1.0 and substantially lower roughness. These category-level priors serve as reference ranges rather than strict physical constraints, thereby retaining instance-specific variations while suppressing inconsistent or atypical artist-defined annotations. This standardized parameterization reduces annotation ambiguity and provides more consistent supervision across heterogeneous data sources.

For the generic \textit{Metal} category, we further provide an optional fine-grained specification to capture the substantial variation in reflective behavior among metallic surfaces. Specifically, metallic materials can be subdivided into four roughness-dependent subclasses: \textit{Mirror Metal}, \textit{Polished Metal}, \textit{Semi-Rough Metal}, and \textit{Matte Metal}. These subclasses correspond to progressively increasing roughness levels, enabling the unified representation to distinguish metallic surfaces with markedly different reflection characteristics while retaining \textit{Metal} as the general semantic category.

To ensure sufficient appearance diversity and geometric coverage, each asset is further rendered from six calibrated viewpoints using the illumination strategy described above.
\end{revision}

\textbf{Data Filtering via Complexity Analysis.}
To ensure the quality and diversity of the dataset, we introduce a complexity metric defined over both albedo and MR representations. The proposed metric jointly captures global photometric variation and local structural details, which are critical for identifying informative training samples. Specifically, given a set of $N$ rendered views $\{I^{(i)}\}_{i=1}^{N}$, we define the overall complexity score as the average across views:
\begin{equation}
C = \frac{1}{N} \sum_{i=1}^{N} \left( 
\lambda_v \, \mathrm{Var}(I^{(i)}) + 
\lambda_g \, \mathbb{E}\left[\|\nabla I^{(i)}\|_2^2\right]
\right).\end{equation} 
where $\mathrm{Var}(\cdot)$ measures the spatial variance of pixel intensities, reflecting global appearance diversity, and $\nabla I$ denotes the image gradient, whose squared $\ell_2$ norm captures high-frequency structural information. The expectation operator $\mathbb{E}[\cdot]$ is computed over all pixels. In practice, we set $\lambda_v = 0.9$ and $\lambda_g = 0.1$ to balance the contributions of photometric variation and structural complexity. Samples with low complexity scores are filtered out, as they typically correspond to trivial textures or overly simple geometry with limited learning signal.
This filtering strategy effectively removes uninformative samples and enhances the overall diversity and representational richness of the dataset, which we find to be crucial for learning richer high-frequency texture details and fine-grained material variations.

\textbf{Iterative Material Recognition and Segmentation.}
The filtered samples are further processed through an iterative vision-language-guided segmentation framework. For assets with reliable artist-defined component decomposition or clear mesh connectivity, we additionally construct part-level segmentation renderings to provide explicit structural cues for material recognition. Specifically, each sufficiently large and visible mesh component is rendered with a unique region ID, while tiny or barely visible components are discarded to reduce recognition noise. For assets with unreliable or overly fragmented component decomposition, we only use the corresponding multi-view renderings as VLM inputs.

Given the available inputs, i.e., multi-view renderings optionally accompanied by ID-based segmentation maps, we employ a vision-language model (VLM) to perform coarse-grained material recognition. When explicit region IDs are available, they help the VLM associate material predictions with specific object parts rather than relying solely on holistic image-level appearance. Conditioned on the predicted material prompts and multi-view observations, we apply SAM3~\cite{carion2025sam} to generate semantic-aware segmentation masks, yielding an initial material-consistent decomposition.

For assets with ID-based segmentation inputs, we further validate the SAM3-predicted masks by matching them with the VLM-assigned component regions using an intersection-over-union (IoU) criterion. Only masks with an IoU greater than 0.7 are retained, ensuring sufficient spatial agreement between part-level material recognition and pixel-level segmentation. For assets without reliable ID-based component maps, where segmentation relies primarily on SAM3 predictions, we instead retain only masks whose SAM3 confidence score exceeds 0.7. These threshold-based filtering steps remove unstable or poorly localized masks and improve the reliability of the resulting material-aware decomposition.

To further improve semantic accuracy, we adopt an iterative refinement strategy. The retained segmented regions are fed back into the VLM for region-level material reassessment, and the updated material predictions are used to guide subsequent segmentation. Finally, the VLM performs a quality assessment on each refined material region, evaluating whether the predicted material label is semantically consistent with the corresponding visual appearance. Only regions with a VLM confidence score above 0.9 are preserved in the final decomposition. This closed-loop process progressively improves material recognition, suppresses unreliable segmentations, and enhances the consistency between semantic material labels and pixel-level masks.

\textbf{Affine Material Projection with Detail Preservation.}
Instead of directly replacing segmented regions, which may overwrite artist-designed details and introduce visible discontinuities, we propose an affine material projection strategy. This operation maps each segmented region into a canonical material domain while preserving its original spatial structure and local material layout. To further retain fine-scale appearance details, we explicitly reintroduce high-frequency residuals extracted from the original renderings using Gaussian filtering:
\begin{equation}
\mathbf{X}_{\mathrm{refined}}
=
\mathbf{X}_{\mathrm{affine}}
+
\lambda_{\mathrm{hf}} \left(\mathbf{X} - G_{\sigma} * \mathbf{X}\right),
\end{equation}
where $\mathbf{X}$ denotes the original rendering or material map, $\mathbf{X}_{\mathrm{affine}}$ denotes the affine-projected result, $G_{\sigma}$ is a Gaussian kernel with standard deviation $\sigma$, $*$ denotes convolution, and $\lambda_{\mathrm{hf}}$ controls the strength of high-frequency residual injection. In our implementation, we set $\lambda_{\mathrm{hf}}=2.0$. This residual injection preserves fine geometric details and realistic surface variations while avoiding the loss of artist-authored high-frequency texture cues.

\textbf{Manual Refinement and Consistency Enforcement.}
Due to the inherent ambiguity in automated parsing—especially under complex structures and cross-view discrepancies—we further conduct a manual inspection and refinement stage. This step ensures semantic correctness and enforces multi-view consistency across all samples in the dataset.

Overall, through the above filtering and cleaning procedures, the pipeline produces a high-quality PBR dataset with consistent material definitions and rich geometric variations. Although the initial data pool is substantially larger, only approximately 20K samples remain suitable for MR training, providing reliable supervision for stable and physically consistent material learning.

\begin{figure*}[t] 
  \centering
  \includegraphics[width=0.95\linewidth]{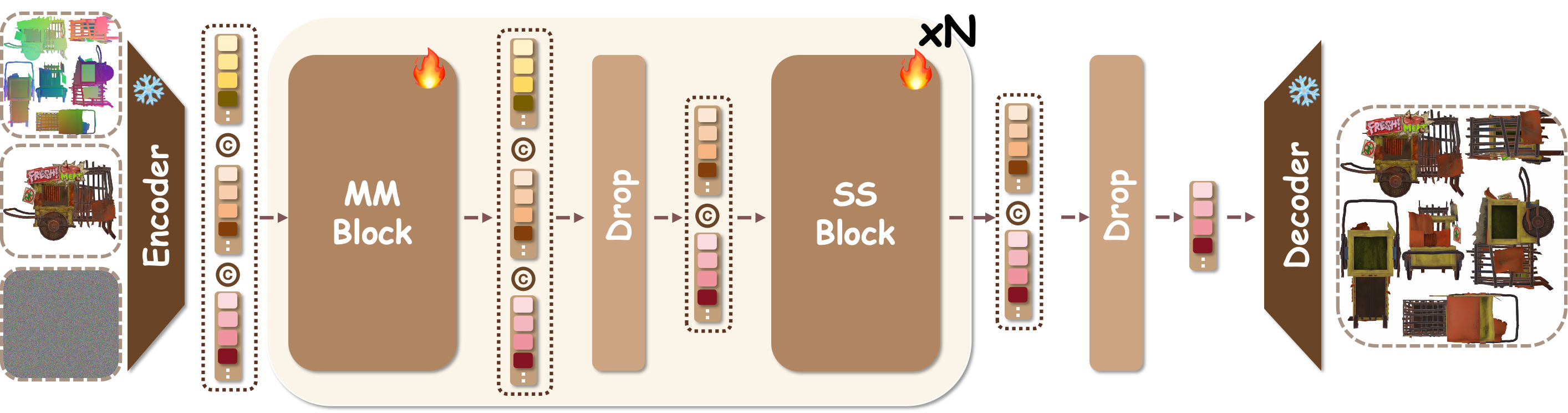} 
  \caption{Overview of our conditional diffusion framework for multi-view albedo generation. Given a noisy input, a reference image, and geometric guidance, the model progressively denoises the input to synthesize consistent multi-view albedo maps. }
  \Description{A pipeline illustrating deterministic diffusion-based multi-view PBR material estimation with packed inputs, conditional guidance, and LoRA fine-tuning.}
  \label{fig:albedo_methods}
\end{figure*}

\section{Multi-View PBR Material Generation}
\rev{After data cleaning, normalization, and refinement, the initial 260K assets are reduced to approximately 20K high-quality PBR-consistent samples, highlighting the scarcity of reliable assets with complete and reliable PBR annotations under the unified specification.} To better exploit this limited supervision, we formulate PBR material generation as a two-stage process: high-resolution multi-view intrinsic decomposition for albedo estimation, followed by albedo-conditioned prediction of full PBR properties.

This design also accounts for the different spatial requirements of PBR components. Since albedo contains rich texture and color details, it is predicted at $1024^2$ resolution, whereas metallic and roughness maps encode relatively smoother reflectance variations and are predicted at $512^2$ resolution. This enables efficient training while preserving high-resolution appearance details and sufficient material fidelity.
\subsection{Illumination-Invariant Albedo Estimation}

\paragraph{\textbf{Dynamic Resolution Permutation-Optimized Packing}}

Motivated by PacTure\cite{fei2025pacture}, multi-view aggregation is formulated as a packing problem to improve spatial efficiency and scalability. Existing fixed-resolution heuristics (e.g., MaxRects) are suboptimal for anisotropic view distributions, often leading to low occupancy and inefficient texture utilization.

To address this, a permutation-based optimization is introduced to jointly determine packing order and implicit per-view resolution. Given rectangles $\mathcal{R}=\{(w_i,h_i)\}_{i=1}^{N}$, permutations $\pi\in S_N$ define sequential insertion under MaxRects with Best Short Side Fit (BSSF), yielding a deterministic layout within a bounding box of area $A_{\mathrm{bbox}}(\pi)=W(\pi)H(\pi)$.

The objective is formulated as maximizing packing efficiency:
\begin{equation}
    \pi^* = \arg\max_{\pi \in S_N} U(\pi)
    = \arg\max_{\pi \in S_N} \frac{\sum_{i=1}^{N} w_i h_i}{W(\pi)H(\pi)}.
\end{equation}

Since exact optimization over the permutation space $S_N$ is intractable due to its factorial complexity, we approximate $\pi^*$ using a Genetic Algorithm (GA). In this framework, each individual encodes a permutation $\pi$, and its fitness is defined by the packing efficiency $U(\pi)$. The population is evolved via rank-based selection, Partially-Mapped Crossover (PMX), and swap mutation operators to explore the combinatorial search space effectively.

To ensure a feasible initialization of the packing process, we estimate the initial bin dimensions $(W_0, H_0)$ based on the total area of all rectangles:
\begin{equation}
    W_0 = \left\lceil \sqrt{\sum_{i=1}^{N} w_i h_i} \right\rceil, \quad
    H_0 = \left\lceil \frac{\sum_{i=1}^{N} w_i h_i}{W_0} \right\rceil.
\end{equation}
If a given permutation results in an infeasible layout under the current bin constraints, we progressively expand the bounding region to ensure successful placement.

Since the input images are first spatially downsampled by the VAE encoder and subsequently tokenized through patchification, we perform multi-view packing directly in the reduced latent-patch space. Specifically, the packing resolution is defined according to the per-view size after a $16\times$ spatial downsampling. This design aligns the packing layout with the effective token resolution used by the network, thereby avoiding unnecessary interference or misalignment among different views after VAE compression and patch partitioning.

Moreover, optimizing the layout in this compact space substantially reduces the computational overhead of the packing procedure. In practice, the complete packing process takes less than 3 seconds per asset, making it efficient enough for large-scale data preprocessing and training.

As illustrated in Fig.~\ref{fig:albedo_methods}, given packed multi-view input $\mathbf{x}_0$, we construct noisy samples via
\begin{equation}
\mathbf{x}_t = \alpha_t \mathbf{x}_0 + (1-\alpha_t) \boldsymbol{\epsilon}, \quad \boldsymbol{\epsilon} \sim \mathcal{N}(0, \mathbf{I}).
\end{equation}

We define the conditioning signal as $\mathbf{c} = \{\mathbf{c}_{geo}, \mathbf{c}_{ref}\}$, where $\mathbf{c}_{geo} = 0.5\,\mathrm{nrm} + 0.5\,\mathrm{pos}$. The reference image $\mathbf{c}_{ref}$ is randomly sampled from a multi-view set $\mathcal{V}$, where each view is rendered either as a shaded PBR image or an albedo image with equal probability.

The velocity predictor $\mathbf{f}_\theta$ is trained using a flow matching objective:
\begin{equation}
\mathcal{L} =
\mathbb{E}_{t, \mathbf{x}_0, \boldsymbol{\epsilon}}
\left[
\left\|
\mathbf{f}_\theta(\mathbf{x}_t, t, \mathbf{c}_{geo},\mathbf{c}_{ref}) - (\mathbf{\epsilon} - \mathbf{x}_t)
\right\|_2^2
\right].
\end{equation}

\paragraph{\textbf{Classifier-free guidance (CFG) conditioned on geometric cues.}}
During inference, we apply classifier-free guidance (CFG) \cite{ho2022classifier}to control the influence of geometric conditioning. 
Given the noisy input $x_t$, the reference condition $\mathbf{c}_{ref}$, and the geometric condition $\mathbf{c}_{geo}$, the guided noise prediction is defined as:
\begin{equation}
\begin{aligned}
\hat{\epsilon}_{\mathrm{cfg}}(x_t, \mathbf{c}_{geo}, \mathbf{c}_{ref}) 
&= f_{\theta}(x_t, \varnothing, \mathbf{c}_{ref}) \\
&\quad + w \cdot \Big(
f_{\theta}(x_t, \mathbf{c}_{geo}, \mathbf{c}_{ref}) 
- f_{\theta}(x_t, \varnothing, \mathbf{c}_{ref})
\Big),
\end{aligned}
\end{equation}
where $w$ denotes the guidance scale and $\varnothing$ represents the dropped geometric condition.

By interpolating between the geometry-conditioned and geometry-unconditioned predictions, the guidance term selectively strengthens structures that are consistent with the input geometry. This encourages the generated material maps to better follow surface boundaries and local geometric variations, while preserving the appearance information provided by the reference condition.

\paragraph{\textbf{Accelerated Inference within Diffusion Distillation Framework}}

To further enhance the practicality of our model, we apply Distribution Matching Distillation (DMD)~\cite{yin2024one} to accelerate inference. Specifically, we distill our multi-step teacher model $f_\eta$ into a few-step student $f_\theta$ by minimizing the reverse KL divergence between their output distributions $p_\theta$ and $p_\eta$:
\begin{equation}
\begin{aligned}
\mathcal{L}_{\mathrm{DMD}}
&= D_{\mathrm{KL}}\left(p_{\theta} \,\|\, p_{\eta}\right) \\
&= \mathbb{E}_{z,t,\epsilon}
\left[
-\left(
\log p_{\eta}(\hat{x})
-
\log p_{\theta}(\hat{x})
\right)
\right].
\end{aligned}
\end{equation}.

Since the densities are intractable, the gradient is estimated via the difference between the teacher and student score functions:

\begin{equation}
\begin{aligned}
\nabla_{\theta} \mathcal{L}_{\mathrm{DMD}}
&=
\mathbb{E}_{z,t,\epsilon}
\left[
-\left(
s_{\eta}(\hat{x})
-
s_{\theta}(\hat{x})
\right)
\nabla_{\theta}\hat{x}
\right].
\end{aligned}
\end{equation}
where $s_\eta$ and $s_\theta$ denote the scores of the teacher and student distributions, respectively, both obtained from their corresponding diffusion models on perturbed samples. With this distillation objective, we transfer the original 30-step CFG-guided sampling process into a compact 6-step inference model, substantially improving generation efficiency while preserving output quality. As a result, the entire model requires only about 40 seconds on average for inference.

\begin{figure}[t] 
  \centering
  \includegraphics[width=\linewidth]{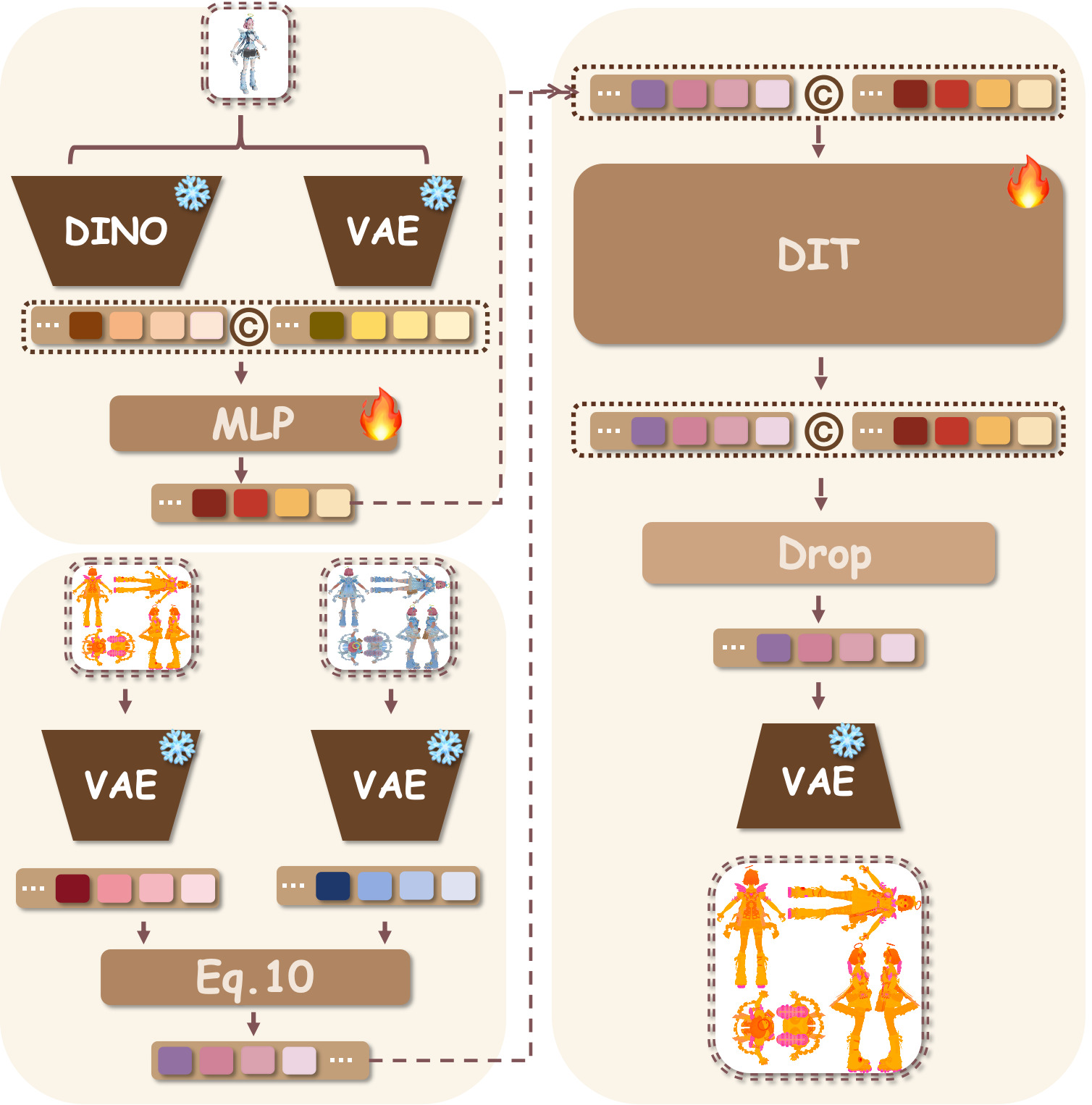} 
  \caption{Overview of our deterministic diffusion framework for multi-view PBR material estimation. Given a packed albedo input $\mathbf{x}_{rgb}$ and a shaded reference image $\mathbf{x}_{ref}$, we formulate a conditional diffusion process to predict the target material representation $\mathbf{x}_{mr}$. }
  \label{fig:pbr_methods}
  \Description{A pipeline illustrating deterministic diffusion-based multi-view PBR material estimation with packed inputs, conditional guidance, and LoRA fine-tuning.}
\end{figure}

\subsection{Intrinsic Material Estimation}
\paragraph{\textbf{Deterministic Inference within Diffusion Framework}}

As illustrated in Fig.~\ref{fig:pbr_methods}, motivated by Lotus2.0~\cite{he2025lotus}, we formulate albedo-to-material prediction as a deterministic inference problem. Under our normalized PBR data definition, each albedo is associated with a unique and physically consistent set of material attributes, making the target PBR representation effectively one-to-one conditioned on the given albedo. Therefore, the mapping is primarily constrained by the normalized material definition and photometric consistency, rather than stochastic variation. Empirically, injecting additional noise during training or inference introduces unnecessary ambiguity and destabilizes optimization. We thus adopt a deterministic prediction formulation for stable and consistent PBR reconstruction.

Accordingly, the problem is formulated as a deterministic conditional mapping within a diffusion framework, where stochasticity is confined to the training-time forward process and inference is noise-free. Under the packed multi-view representation, given albedo $\mathbf{x}_{rgb}$, shaded reference $\mathbf{x}_{ref}$, and target material $\mathbf{x}_{mr}$, the diffusion pair is defined as:
\begin{equation}
\mathbf{x}_t = \alpha_t \mathbf{x}_{mr} + \sigma_t \mathbf{x}_{rgb}.
\end{equation}

The reference image $\mathbf{x}_{ref}$ is encoded using a hybrid feature representation that jointly captures semantic structure and low-level appearance information. 
Specifically, we employ a pretrained DINO encoder~\cite{oquab2023dinov2} to extract high-level semantic features, which provide robust global cues that are relatively invariant to viewpoint and local appearance variations. 
In parallel, a VAE encoder~\cite{kingma2013auto} is used to obtain latent appearance features that preserve fine-grained photometric and material-related details:
\begin{equation}
\mathbf{f}_{dino} = \mathcal{E}_{dino}(\mathbf{x}_{ref}), 
\quad
\mathbf{f}_{vae} = \mathcal{E}_{vae}(\mathbf{x}_{ref}).
\end{equation}

The resulting semantic and appearance embeddings are concatenated and fused through an MLP to produce the reference conditioning representation:
\begin{equation}
\mathbf{z}_{ref} = \mathrm{MLP}\left([\mathbf{f}_{dino}, \mathbf{f}_{vae}]\right).
\end{equation}

Given the reference condition $\mathbf{z}_{ref}$, the model is trained to predict the residual between the RGB observation and the target metallic--roughness representation. 
The training objective is formulated as an $L_2$ reconstruction loss:
\begin{equation}
\mathcal{L} =
\mathbb{E}_{t, \mathbf{x}_{mr}}
\left[
\left\|
f_\theta(\mathbf{x}_t, \mathbf{z}_{ref}) 
-
(\mathbf{x}_{rgb}-\mathbf{x}_{mr})
\right\|_2^2
\right].
\end{equation}
\begin{figure*}[!t]
  \centering
  \includegraphics[width=0.95\textwidth]{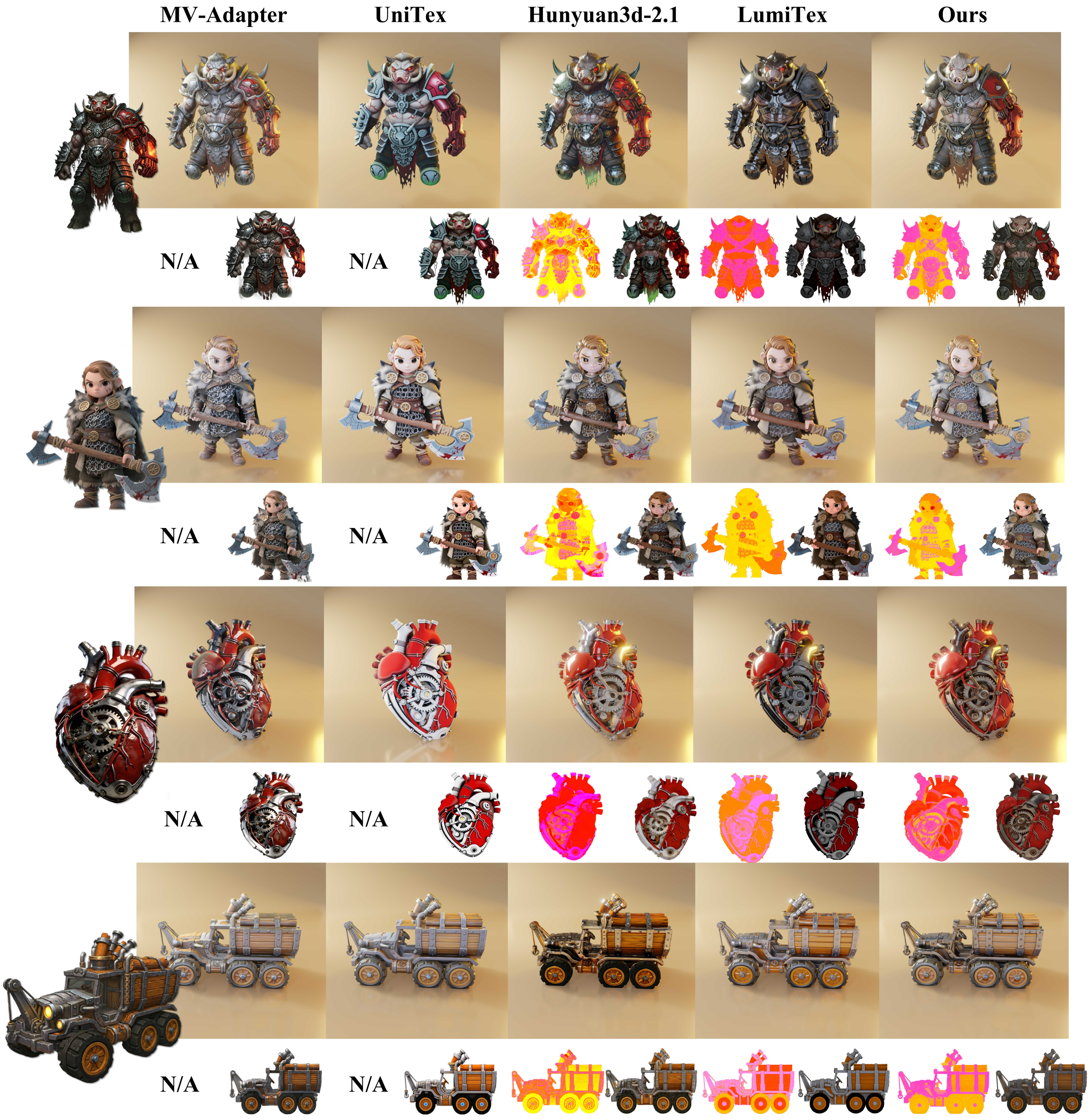}
  \caption{
    Qualitative comparison of texture and PBR material generation across different methods. Each example is arranged in two rows: the first row shows the rendered results under a fixed illumination setting, while the second row presents the predicted material maps, including the MR map, i.e., metallic and roughness channels, and the generated albedo texture. Compared with existing methods, our approach produces sharper and more detailed albedo textures, while also yielding cleaner, more spatially coherent, and semantically consistent MR predictions across diverse samples.
  }
  \label{fig:compare}
\end{figure*}

\section{Implementation Details}
\textbf{Training.} We adopt a parameter-efficient fine-tuning strategy by freezing the pretrained backbone and optimizing only the LoRA parameters~\cite{hu2022lora}. For both the albedo estimation and PBR material generation stages, the LoRA rank and scaling factor are set to 256. Our model is built upon the FLUX.2 backbone~\cite{flux-2-2025} and trained on 32 NVIDIA A800 GPUs with a total batch size of 64. We use the AdamW optimizer with $\beta_1 = 0.9$ and $\beta_2 = 0.999$, a learning rate of $5 \times 10^{-5}$, and a linear warm-up schedule over the first 2,000 optimization steps.

For timestep sampling, we adopt a logit-normal sampling strategy to balance training across different noise levels. In all experiments, the mean and standard deviation of the logit-normal distribution are set to $\mu=-1$ and $\sigma=1$, respectively. This setting is used consistently for packed multi-view inputs during training. To improve training stability and support high-resolution albedo estimation, we employ a hierarchical resolution training strategy for the albedo stage. Specifically, the albedo model is first trained at $512 \times 512$ resolution for 5,000 steps to learn stable global appearance, coarse spatial layout, and cross-view consistency. Starting from this checkpoint, we further fine-tune it at $1024 \times 1024$ resolution for an additional 10,000 steps to enhance high-frequency texture details, boundary fidelity, and fine-scale appearance variations.

In contrast, the PBR material generation stage focuses on metallic--roughness (MR) prediction, whose spatial variations are generally smoother and do not require the same resolution as albedo. Therefore, the MR model is trained only at $512 \times 512$ resolution for 15,000 steps, reducing computational cost while preserving sufficient material fidelity.

\begin{figure*}[htbp]
  \centering
  \includegraphics[width=0.95\textwidth]{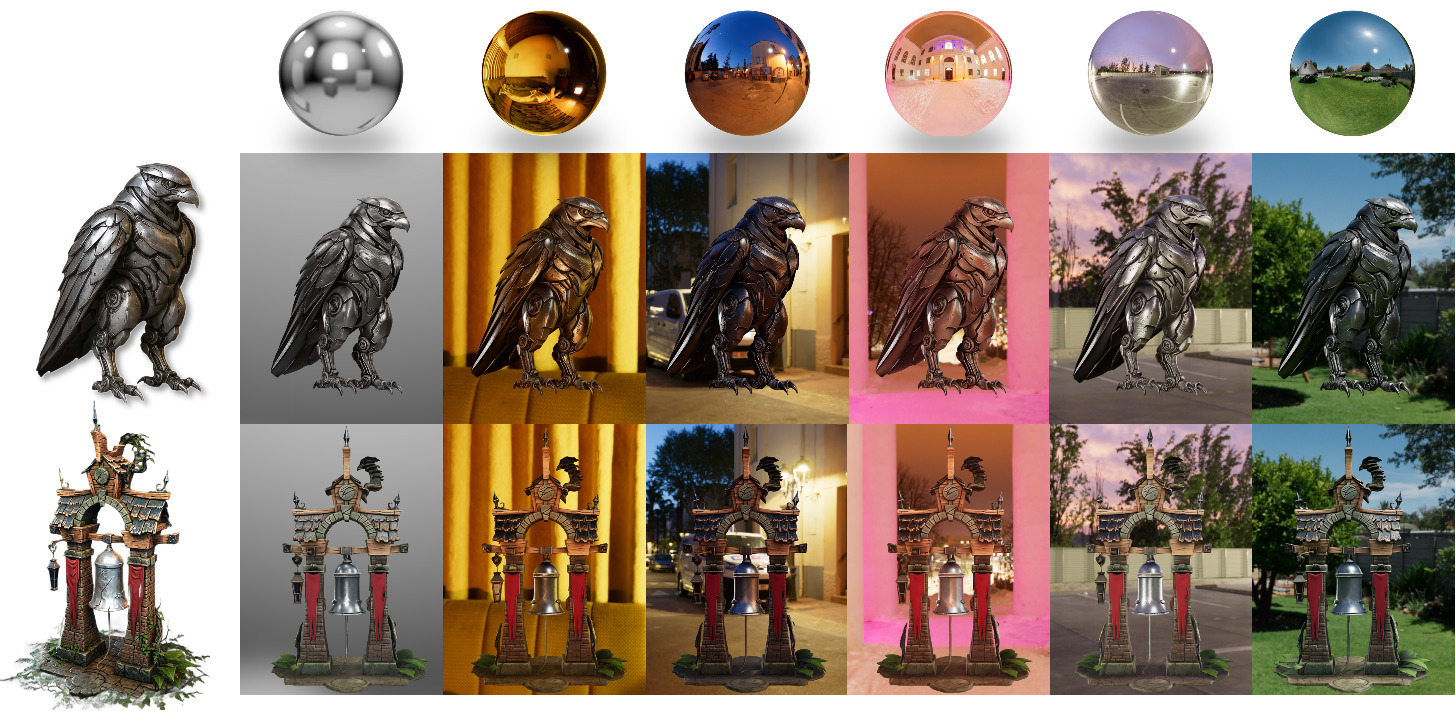}
  \caption{Visualization results of PBR materials and texture maps generated for two given meshes under image guidance. The rendered results under diverse lighting environments demonstrate that the model effectively adapts to varying illumination conditions. In particular, metallic regions exhibit physically plausible reflections, accurately capturing environmental lighting and producing realistic appearance.}
  \Description{A pipeline illustrating deterministic diffusion-based multi-view PBR material estimation with packed inputs, conditional guidance, and LoRA fine-tuning.}
  \label{fig:light_varitions}
\end{figure*}

\begin{figure*}[ht]
  \centering
  \includegraphics[width=\textwidth]{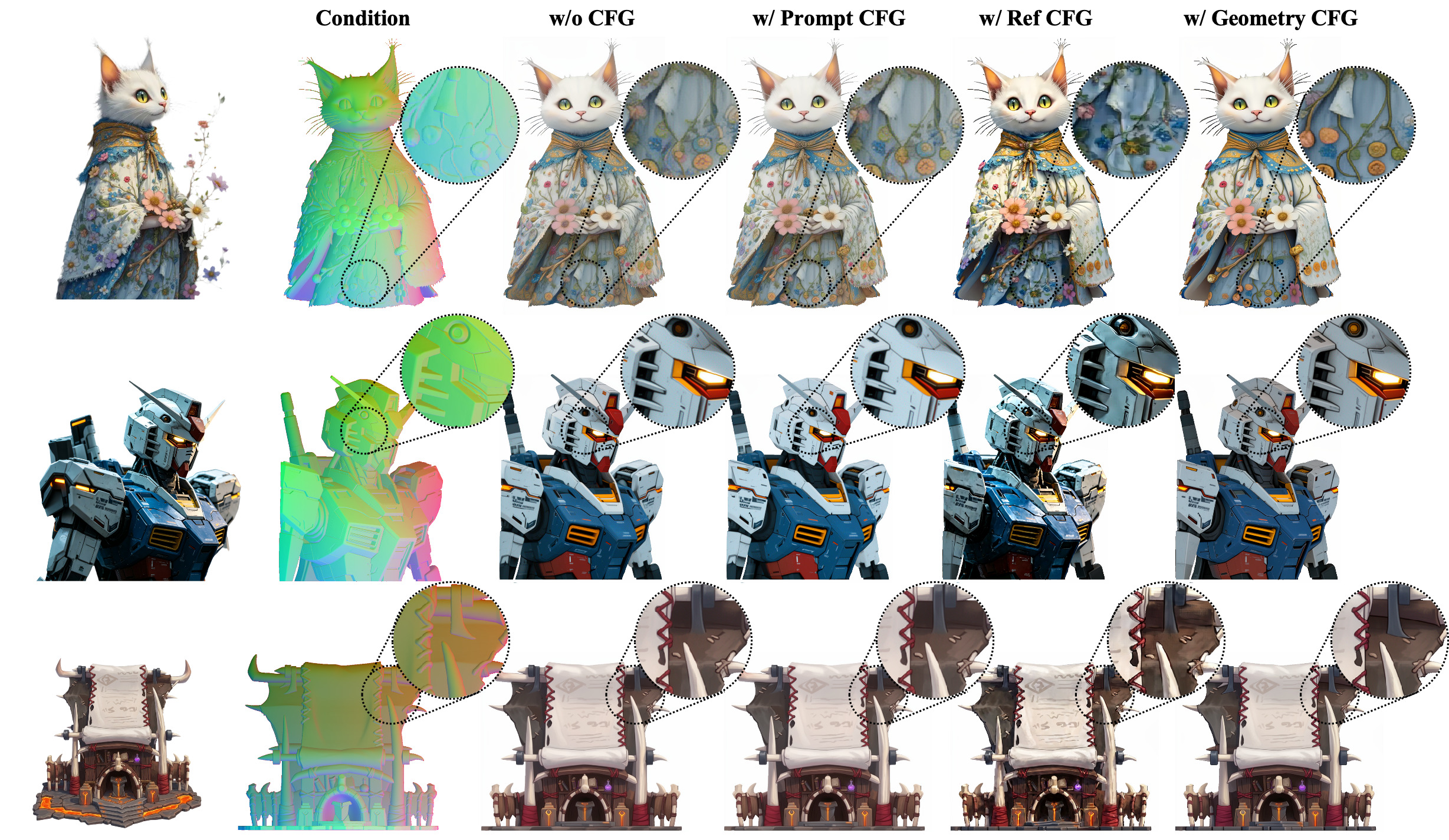}
  \caption{Qualitative comparison of texture generation using geometry-based classifier-free guidance (CFG) versus alternative CFG strategies. The results show that geometry-guided CFG produces sharper textures that more closely align with the conditioning image, leading to improved structural fidelity and overall visual consistency.}
  \Description{A pipeline illustrating deterministic diffusion-based multi-view PBR material estimation with packed inputs, conditional guidance, and LoRA fine-tuning.}
  \label{fig:ablation1}
\end{figure*}

\begin{figure}[ht]
  \centering
  \includegraphics[width=1.0\linewidth]{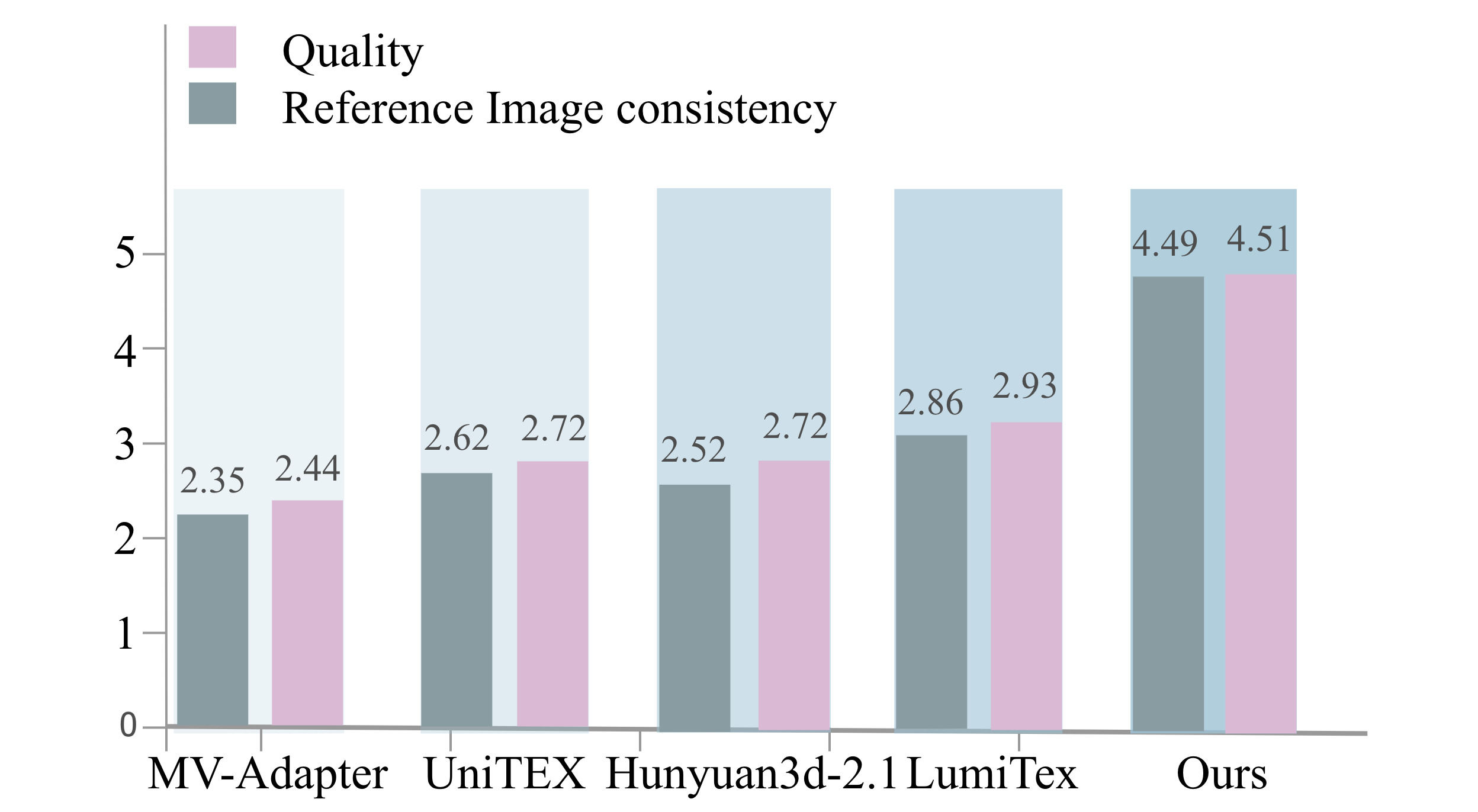}
    \caption{User-study results comparing perceptual quality and reference-image consistency across different methods for multi-view image generation. The bar chart reports the average scores for each method, with higher values indicating better perceived quality and consistency with the reference image.}
  \Description{A pipeline illustrating deterministic diffusion-based multi-view PBR material estimation with packed inputs, conditional guidance, and LoRA fine-tuning.}
  \label{fig:method_study_comapre}
\end{figure}

\begin{figure}[ht]
  \centering
  \includegraphics[width=1.0\linewidth]{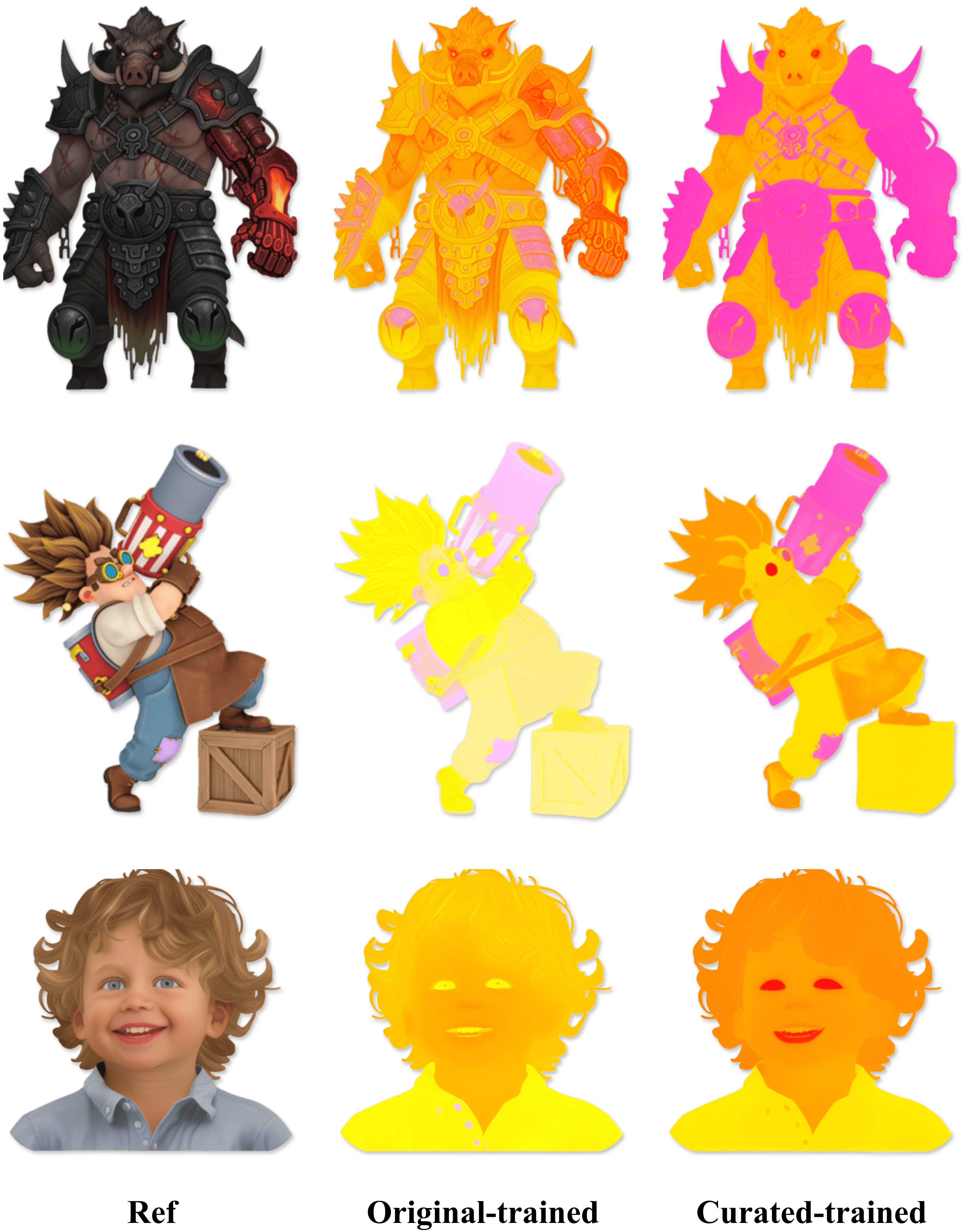}
    \caption{Visual comparison between models trained on original artist-defined PBR data and our unified normalized PBR data. The first column shows the reference image, the second column presents results from the model trained on the original artist-defined PBR data, and the third column shows results from the model trained on our unified normalized PBR data.}
  \label{fig:pbr_ablation}
  \Description{A pipeline illustrating deterministic diffusion-based multi-view PBR material estimation with packed inputs, conditional guidance, and LoRA fine-tuning.}
\end{figure}

\section{Experiments}
To systematically evaluate the performance of our framework, we conduct a comprehensive set of experiments focusing on both the quality and efficiency of PBR generation. We perform rigorous quantitative and qualitative comparisons against state-of-the-art methods in PBR material estimation and generation.
\subsection{Qualitative Results}

\paragraph{\textbf{Visual Comparisons against Existing Methods.}} We evaluate single-view texture generation fidelity under arbitrary novel viewpoints. 
To this end, we construct a challenging benchmark from Objaverse~\cite{deitke2023objaverse}, consisting of 200 high-quality 3D assets with diverse geometries and complex surface structures. 
For each asset, we render six random viewpoints for both quantitative and qualitative evaluation of the generated texture and material maps.

As shown in Fig.~\ref{fig:compare}, we compare our method with representative state-of-the-art approaches. 
MvAdapter~\cite{huang2025mv} often fails to disentangle intrinsic albedo from illumination, resulting in residual shading artifacts. 
Unitex~\cite{liang2025unitex} suppresses lighting effects to some extent but does not explicitly recover full PBR material properties. 
\rev{Hunyuan3d-2.1~\cite{hunyuan3d2025hunyuan3d}} predicts both albedo and material maps, yet its material predictions are often inconsistent across geometrically or semantically similar regions. 
Although LumiTex~\cite{bao2025lumitex} produces visually plausible textures, its material estimates remain unstable and may lack cross-view physical consistency. 
Moreover, these methods tend to lose fine-grained albedo details compared with the input image, leading to reduced local texture fidelity.

In contrast, our method recovers albedo maps with richer details and stronger alignment with the input appearance, while effectively reducing illumination contamination. 
It also generates more stable and spatially consistent PBR material maps whose properties are better correlated with the recovered albedo and remain coherent across viewpoints, demonstrating improved texture fidelity, cross-view consistency, and physical plausibility.

\paragraph{\textbf{Visual Results under Varying Illumination Conditions.}} As illustrated in Fig.~\ref{fig:light_varitions}, we present several representative cases rendered under a diverse set of environmental lighting conditions. The results show that the estimated PBR material properties effectively capture intrinsic surface characteristics and exhibit appropriate responses to illumination changes. In particular, the rendered appearances vary consistently with the incident lighting, demonstrating coherent shading, specular behavior, and reflectance patterns across different environments. This indicates that our method produces physically plausible and lighting-aware material representations, enabling stable and realistic appearance synthesis under novel illumination conditions.

\begin{table*}[t]
\caption{Quantitative Comparison with State-of-the-Art Methods on PBR Texture Generation Quality and Efficiency.}
\label{tab:comparison}
\centering
\small
\begin{tabular}{lcccccccc}
\toprule
Method & FID $\downarrow$ & CMMD $\downarrow$ & LPIPS $\downarrow$ & CLIP-I $\uparrow$ & CLIP-FID $\downarrow$ & Inference Time (s) $\downarrow$  & Albedo Resolution $\uparrow$ & MR Resolution $\uparrow$ \\
\midrule
MvAdapter     & 165.49 & 2.2235 & 0.2383 & 0.9097 & 22.911 & \textbf{36} & 768 & None \\
UniTex        & 161.27 & 2.1898 & 0.2174 & 0.9052 & 22.623 & 220 & 512 & None\\
\rev{Hunyuan3d-2.1}   & 154.88 & 2.2625 & 0.2167 & 0.9132 & 20.132 & 48 & 768 & \textbf{768} \\
LumiTex       & 144.61 &1.8677 & 0.1992 & 0.9215 & 16.182 & 110+110 & 512 & 512 \\
\textbf{Ours} & \textbf{121.25} & \textbf{1.4652} & \textbf{0.1612} & \textbf{0.9411}& \textbf{13.358} & 40+12 & \textbf{1024}  & 512\\ 
\bottomrule
\end{tabular}
\end{table*}

\subsection{Quantitative Analysis}

To quantitatively evaluate the texture fidelity of the generated assets, we adopt a comprehensive set of metrics covering distribution-level realism, perceptual quality, and semantic alignment. As shown in Tab.~\ref{tab:comparison}, our method consistently outperforms existing approaches across all evaluated criteria. Specifically, CMMD~\cite{jayasumana2024rethinking} and FID~\cite{heusel2017gans} evaluate the distributional fidelity of generated textures, while LPIPS~\cite{zhang2018unreasonable} measures perceptual similarity to reference appearances. CLIP Score~\cite{hessel2021clipscore} and CLIP-FID~\cite{parmar2022aliased} further assess semantic alignment and feature-space consistency. Overall, these metrics requires characterize the visual and semantic quality of the generated texture appearance. These results demonstrate that our method produces visually faithful textures that are well aligned with the input guidance.

Furthermore, we conduct a user study to compare our method with existing approaches, as shown in Fig.~\ref{fig:method_study_comapre}. The results further confirm the effectiveness of our framework, where our method achieves state-of-the-art performance in terms of visual quality, material realism, and consistency with the input reference.

\subsection{Ablation Study}
\paragraph{\textbf{Effect of classifier-free guidance (CFG) conditioned on geometric cues.} }To evaluate the effectiveness of classifier-free guidance (CFG) conditioned on geometric cues, we train and compare several CFG variants under different conditioning strategies, as illustrated in Fig.~\ref{fig:ablation1}. We first observe that conventional text-based CFG provides negligible improvement when strong image conditioning is present, with minimal changes in the generated results. This suggests that the model tends to prioritize image tokens over textual guidance during cross-attention, leading to an implicit suppression of text-conditioned signals. In contrast, CFG based on reference images does not lead to consistent quality improvements and may introduce visual artifacts, indicating potential conflicts between image-level guidance and the primary conditioning signal. By comparison, CFG conditioned on geometric cues consistently improves structural fidelity and fine-grained detail consistency. This indicates that geometric information provides a more stable and complementary conditioning signal, effectively constraining the generation process and improving alignment between the synthesized appearance and the underlying geometry.
Furthermore, as shown in Tab.~\ref{tab:cfg_ablation}, we quantitatively evaluate the effect of the proposed geometry-conditioned CFG by comparing model performance with and without this guidance strategy. The results demonstrate that enabling geometry-conditioned CFG substantially improves performance across evaluation metrics, further confirming its effectiveness in enhancing generation quality and geometry-aware consistency.

\begin{table}[t]
\caption{Ablation study on the effect of classifier-free guidance (CFG) in our framework. Enabling CFG significantly improves generation quality and semantic consistency across all evaluation metrics.}
\label{tab:cfg_ablation}
\centering
\small
\setlength{\tabcolsep}{4pt}
\begin{tabular}{lcccccc}
\toprule
Method & Setting & FID $\downarrow$ & CMMD $\downarrow$ & LPIPS $\downarrow$ & CLIP-I $\uparrow$  & CLIP-FID $\downarrow$ \\
\midrule
\multirow{2}{*}{Ours}
& w/o CFG & 128.55 & 1.5439 & 0.1688 & 0.9342 & 15.172 \\
& w/  CFG  & \textbf{121.25} & \textbf{1.4652} & \textbf{0.1612} & \textbf{0.9411} &  \textbf{13.358} \\
\bottomrule
\end{tabular}
\end{table}

\begin{figure*}[t]
  \centering
  \includegraphics[width=0.95\textwidth]{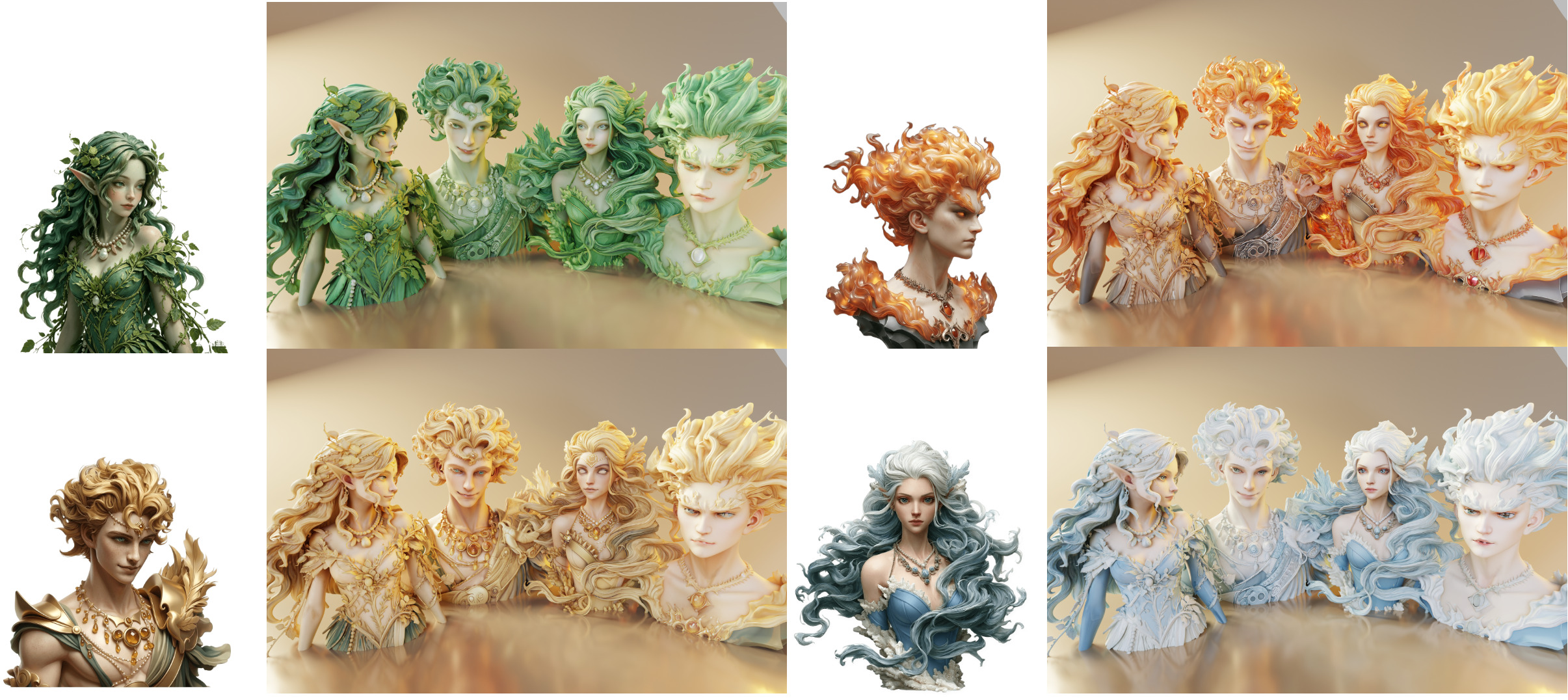}
  \caption{Visualization results of arbitrary image-guided texture generation and mesh stylization. The leftmost images show the reference guidance images, while the rendered results on the right are obtained by generating textures conditioned on each reference image and applying them back onto the same underlying mesh. Despite the substantial appearance differences among the guidance images, our method produces coherent and harmonious stylized textures that are well integrated with the shared geometry, demonstrating strong image-guided stylization capability.}
  \label{fig:vis_style}
\end{figure*}

\begin{figure}[ht]
  \centering
  \includegraphics[width=0.95\linewidth]{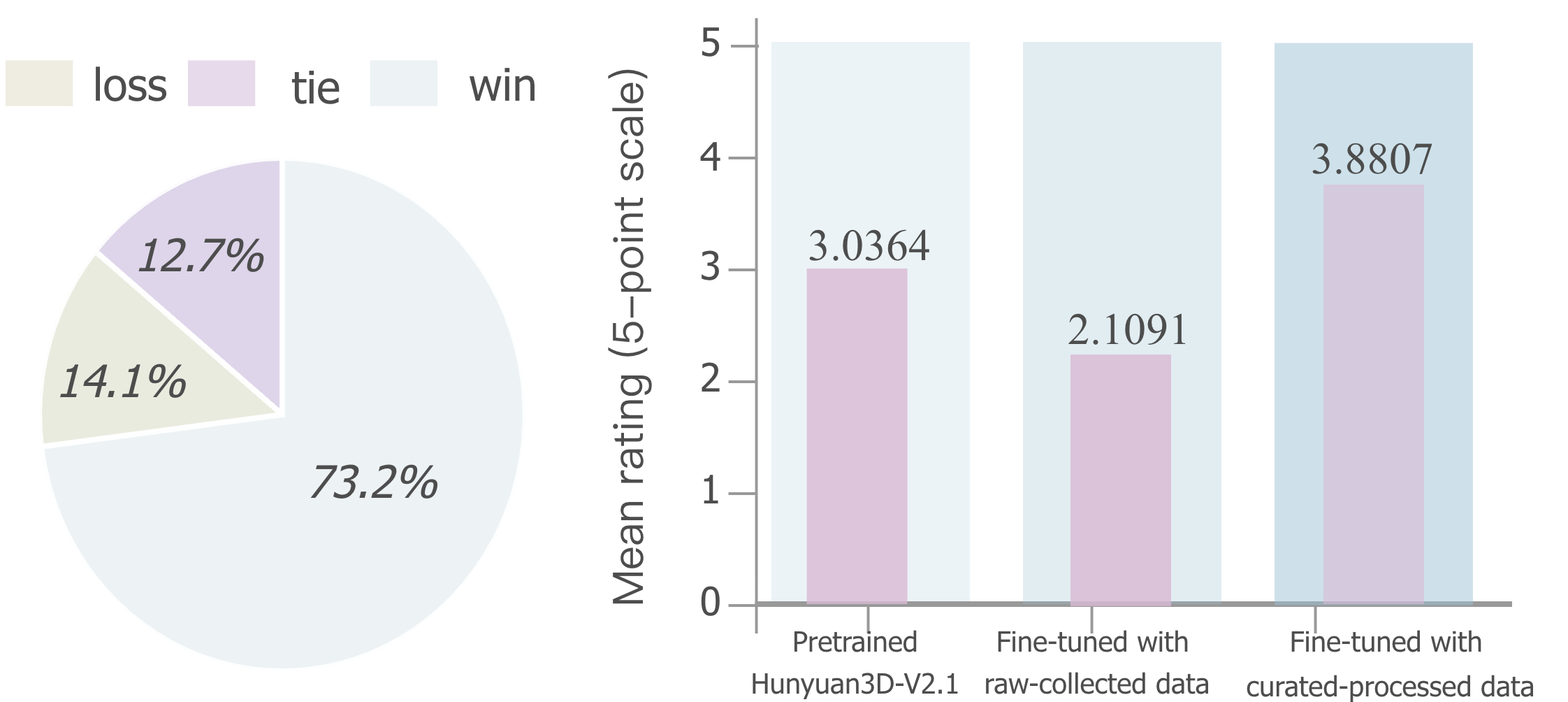}
\caption{\rev{{User study on perceptual PBR material quality.}
\textit{Left:} Pairwise preferences between our model trained on the curated, standardized annotations and that trained on the original annotations, where win, tie, and loss are defined from the perspective of the former.
\textit{Right:} Mean five-point ratings of Hunyuan3D-V2.1 under the pretrained setting and after fine-tuning on either the raw collected data or our curated data.}}
  \Description{A pipeline illustrating deterministic diffusion-based multi-view PBR material estimation with packed inputs, conditional guidance, and LoRA fine-tuning.}
  \label{fig:pbr_user_study}
\end{figure}

\begin{figure*}[t]
  \centering
  \includegraphics[width=0.95\textwidth]{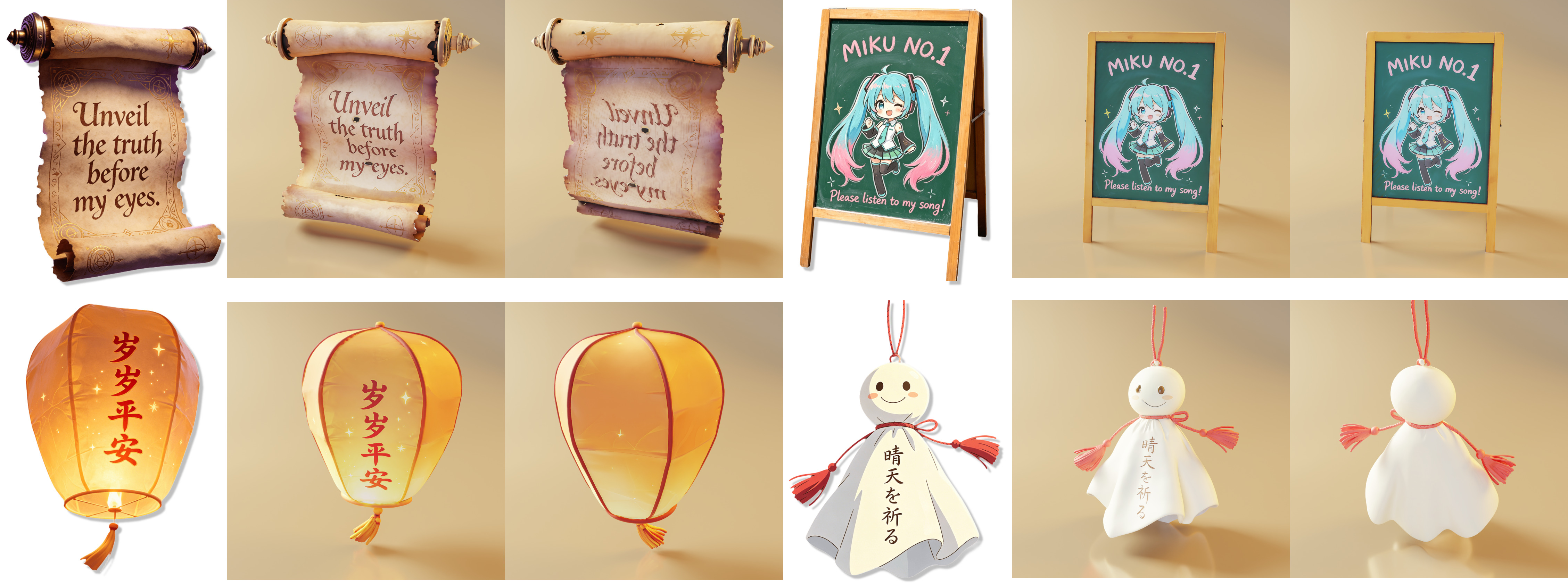}
  \caption{Visualization of text and logo fidelity under reference guidance. For each example, the leftmost image denotes the reference input containing distinctive textual or logo-like patterns. The two rendered results on the right show the textured mesh after back-projecting the generated maps, where the left rendering presents the frontal view and the right rendering presents the back view. The results demonstrate that our method can preserve fine-grained text and logo structures from the reference image while maintaining coherent appearance across different viewpoints.}
  \label{fig:text_logo}
\end{figure*}

\begin{figure*}[htbp]
  \centering
  \includegraphics[width=0.95\textwidth]{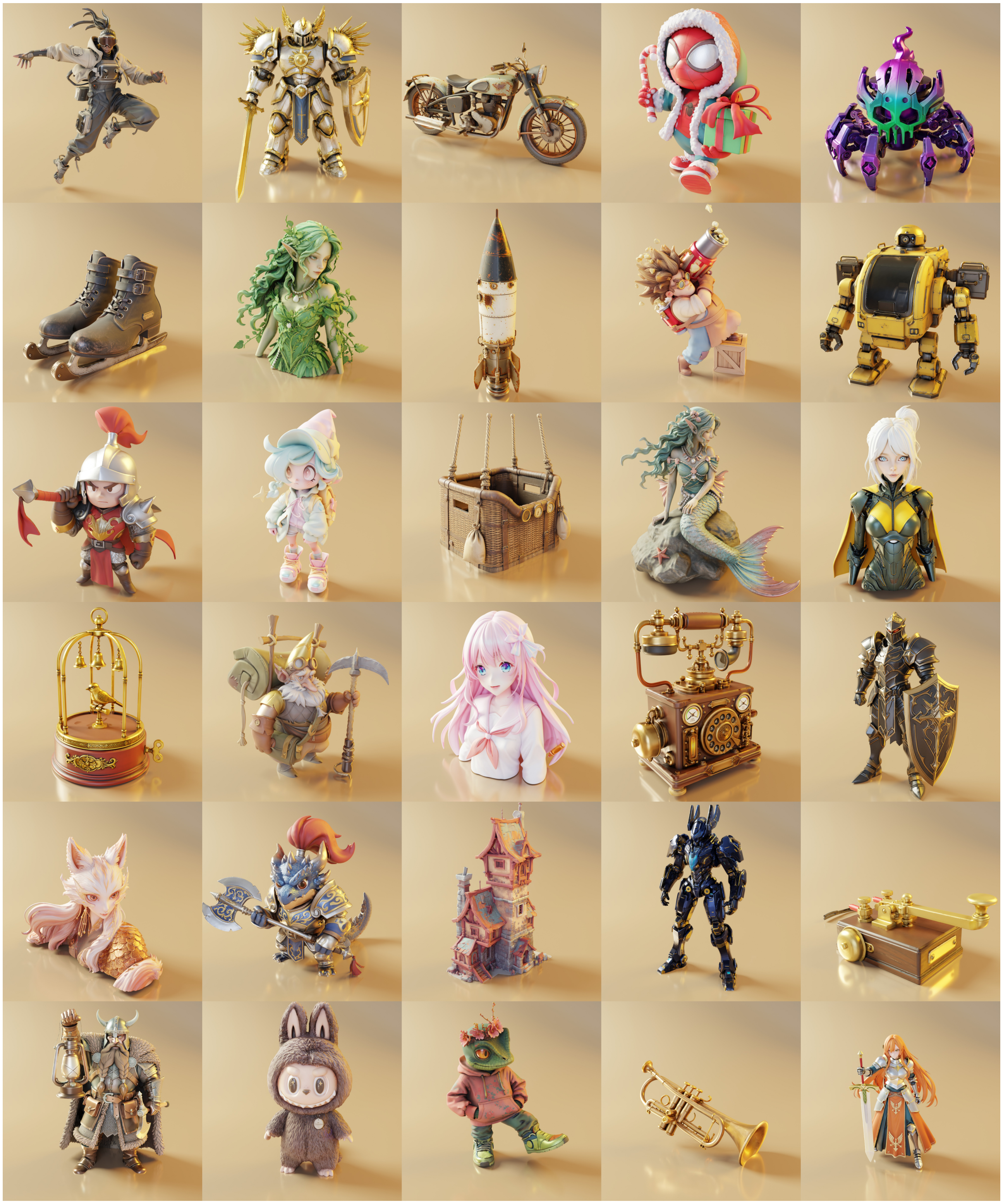}
  \caption{Additional qualitative results on diverse inputs. Our method demonstrates strong generalization across a wide range of mesh geometries, styles, and input variations, consistently producing high-quality textures along with semantically accurate and physically plausible PBR materials.}
  \label{fig:supp_vis}
\end{figure*}

\paragraph{\textbf{Effect of PBR Material Consistency on Inference Stability.}}
To evaluate the impact of unified PBR material specifications, we conduct a controlled ablation study using the same curated dataset under two settings: training with the original artist-defined material parameters and training with our standardized material specification. As shown in Fig.~\ref{fig:pbr_ablation}, the model trained with unified materials produces more stable and consistent MR predictions across regions sharing the same semantic material, whereas the model trained on artist-defined parameters exhibits visible inconsistencies caused by non-uniform material definitions.

\begin{revision}
    
Because our standardization modifies the original PBR annotations, evaluation against either the original or standardized labels would introduce bias. We therefore conduct two complementary user studies to assess perceptual material quality. First, we perform a controlled ablation on our model, varying only whether PBR standardization is applied. Participants compare paired mesh renderings of the predicted MR attributes in terms of material consistency and physical plausibility. As shown in Fig.~\ref{fig:pbr_user_study} (left), our standardized variant achieves a $73.2\%$ win rate, with $14.1\%$ ties and $12.7\%$ losses. Second, we fine-tune Hunyuan3D-V2.1 \cite{hunyuan3d2025hunyuan3d} using either the raw collected data or our curated data and evaluate the results on a five-point scale. As shown in Fig.~\ref{fig:pbr_user_study} (right), fine-tuning on our curated data achieves the highest mean rating of $3.88$, compared with $3.04$ for the pretrained model and $2.11$ for fine-tuning on the raw data. These results demonstrate that our unified PBR specification reduces supervision ambiguity and improves material consistency and physical plausibility across different model architectures.

\end{revision}
\begin{revision}
\paragraph{\textbf{Evaluation of PBR Stability under Semantic-Preserving Style Variations.}}

Beyond improving consistency across regions with identical semantic materials, we further evaluate whether our standardized material representation improves the physical stability of predictions under appearance variations. Ideally, inputs depicting the same object with different color distributions or visual styles should produce similar PBR properties, since their underlying physical materials remain unchanged.

To evaluate this property, we generate multiple semantic-preserving style variants of the same input image using GPT-based image editing. We then predict PBR maps for each variant and measure the consistency between predictions using the mean squared error (MSE).

\begin{table}[t]
\caption{\rev{Ablation study on the effect of PBR material standardization. The standardized material specification improves prediction consistency under semantic-preserving appearance variations.}}
\label{tab:pbr_stability}
\centering
\small
\setlength{\tabcolsep}{4pt}

\ifshowrevision
    \color{revisionblue}
\fi

\begin{tabular}{lc}
\toprule
Training Data & PBR Consistency MSE $\downarrow$ \\
\midrule
Artist-defined Materials & 0.0513 \\
Standardized Materials & \textbf{0.0258} \\
\bottomrule
\end{tabular}

\end{table}

As show in Tab. \ref{tab:pbr_stability}, the curated dataset achieves significantly lower prediction variance, indicating that our normalization strategy improves PBR stability and reduces ambiguity introduced by inconsistent material annotations. Importantly, our method does not enforce identical PBR values for all instances within the same semantic category. Instead, it regularizes inconsistent material definitions while preserving spatial variations and high-frequency appearance details.

\end{revision}


\subsection{Additional Applications and Analysis}

\paragraph{Texture Stylization from Arbitrary Images}
Our model supports stylized texture generation from arbitrary input images without requiring domain-specific constraints or curated references. Given an input image with diverse appearance characteristics, the model extracts high-level visual cues and translates them into coherent texture representations suitable for 3D assets. This flexibility enables a wide range of applications, including rapid texture prototyping, image-guided mesh stylization, and content creation from in-the-wild images. In particular, the generated textures preserve the overall stylistic attributes of the input while maintaining clean appearance, coherent surface details, and stable visual quality under varying illumination conditions. As shown in Fig.~\ref{fig:vis_style}, our method produces high-quality stylized textures that can be directly applied to 3D meshes, resulting in visually realistic renderings with strong detail fidelity, harmonious style transfer, and consistent appearance across different scenarios.

\paragraph{{Text and Logo Fidelity under Reference Guidance.}}
In addition to overall appearance and material consistency, preserving distinctive local patterns such as text, logos, and graphic symbols is also important for reference-guided texture generation. These elements often contain high-frequency structures and precise semantic layouts, making them challenging to reproduce faithfully on complex 3D surfaces. As shown in Fig.~\ref{fig:text_logo}, our method effectively transfers such reference-specific details onto the target mesh while maintaining clear boundaries, recognizable character shapes, and coherent spatial alignment. The results demonstrate that our model can preserve fine-grained textual and logo-like patterns under reference guidance, further improving the fidelity and controllability of the generated textures.

\paragraph{Extended Qualitative Results}

As shown in Fig.~\ref{fig:supp_vis}, we present additional results on a diverse set of inputs, where the generated texture and PBR material maps are applied back onto the corresponding meshes for rendering under novel illumination conditions. Our method consistently produces high-quality textures and physically plausible material properties. The generated albedo maps are clean and largely free of shading artifacts, while the recovered reflectance parameters exhibit strong spatial coherence.


\section{Limitations and Future Work}

Although our data construction pipeline improves the consistency and reliability of PBR material annotations, it still involves limited manual intervention. Specifically, the final validation stage relies on human inspection to identify ambiguous, low-quality, or visually implausible samples. This process helps maintain dataset quality but may introduce subjective biases, as different annotators may have varying judgments regarding material plausibility, texture fidelity, and visual consistency. Moreover, manual screening increases the cost of dataset construction and may affect scalability to larger asset collections.

\begin{revision}
Furthermore, our current approach focuses on achieving relative physical consistency rather than strict physical correctness. Since the material definitions are derived from artist-created assets, the resulting PBR distributions inevitably contain certain artistic approximations and may deviate from the true physical properties of real-world materials. While our standardization strategy reduces annotation ambiguity and improves consistency under a unified rendering protocol, it does not fully recover physically measured material properties. In future work, we plan to incorporate physically acquired material measurements, such as captured SVBRDFs from dedicated acquisition devices, to develop more physically accurate and grounded material representations.
\end{revision}

\section{Conclusion}

In this work, we investigate the fundamental challenges of learning physically based material properties from image observations, with particular emphasis on the ambiguities arising from inconsistent illumination conditions and non-uniform PBR material definitions. We show that variations in environment lighting and artist-defined material parameterizations can substantially entangle intrinsic material attributes with view- and illumination-dependent shading effects, resulting in unstable supervision signals and inconsistent material predictions. To mitigate these issues, we establish a controlled data generation pipeline with consistent illumination and introduce a unified specification for PBR material properties. Built upon this \rev{normalized PBR representation informed by empirical material priors}, we further train a state-of-the-art generative model capable of synthesizing high-quality texture maps together with coherent PBR material properties. Through extensive experiments, including ablation studies on material consistency, we demonstrate that our data-centric design effectively reduces ambiguity in the learning process and improves the stability, coherence, and physical plausibility of model inference, particularly across spatially separated regions that share the same underlying material.

Overall, our findings underscore the critical role of data consistency in disentangling intrinsic material properties from image observations. Beyond achieving strong generation performance, this study provides practical insights for constructing robust, scalable, and \rev{more consistent and empirically grounded PBR material generation systems.}
\section{Acknowledgments}
This work was supported by National Natural Science Foundation of China (62472213), Gusu Innovation \& Entrepreneurship Leading Talents Program (ZXL2024361), and computing resources from DreamTech.

\FloatBarrier
\newpage
\bibliographystyle{ACM-Reference-Format}
\bibliography{references}



\end{document}